\documentclass[11pt]{article}

\PassOptionsToPackage{hyperfootnotes=false}{hyperref}
\usepackage[preprint]{acl}
\usepackage{times}
\usepackage{latexsym}
\usepackage[T1]{fontenc}
\usepackage[utf8]{inputenc}
\usepackage{microtype}
\usepackage{inconsolata}
\usepackage{graphicx}
\usepackage{booktabs}
\usepackage{amsmath}
\usepackage{amssymb}
\usepackage{multirow}
\usepackage{subcaption}
\usepackage{adjustbox}
\usepackage{placeins}
\usepackage[most]{tcolorbox}
\usepackage{xspace}
\usepackage{float}
\usepackage{url}

\newcommand{\method}{\textsc{AlphaPilot}\xspace}
\newtcblisting[auto counter, number within=section]{PromptBlock}[2][]{
  enhanced,
  breakable,
  listing only,
  listing engine=listings,
  colback=black!2,
  colframe=black!55,
  colbacktitle=black!6,
  coltitle=black,
  title={#2},
  fonttitle=\bfseries,
  boxrule=0.5pt,
  arc=0pt,
  left=0.8mm,
  right=0.8mm,
  top=0.5mm,
  bottom=0.5mm,
  before skip=4pt,
  after skip=4pt,
  listing options={
    basicstyle=\scriptsize\ttfamily,
    columns=fullflexible,
    keepspaces=true,
    breaklines=true,
    showstringspaces=false,
    aboveskip=0pt,
    belowskip=0pt
  },
  #1
}

\title{AlphaPilot: Multi-Turn Tool-Using LLM Agents for Adaptive Alpha Mining with Reinforcement Learning}

\author{Peiqi Jiang \quad Fan Zhang \quad Xin Xu \quad Tianhao Chen \quad Can Yang\thanks{Corresponding author.} \\
  Department of Mathematics, Hong Kong University of Science and Technology}

\begin{document}
\maketitle

\begin{abstract}
Formulaic alpha mining seeks interpretable expressions that rank assets by
future returns. The task is challenging because it demands simultaneous
processing of unstructured market descriptions, historical data, and
quantitative evaluation of candidate factors, yet most existing methods exploit
only a subset of these signals. 
We introduce \method, a reinforcement-learning framework that trains multi-turn tool-using LLM agents for adaptive alpha mining.
The agent conditions on market descriptions, proposes factor expressions, queries real-data evaluator feedback, and updates a retained factor pool across turns.
This design jointly leverages market text, LLM financial knowledge, real data, and evaluation feedback.
At inference time, \method performs day-by-day search and pool updates to
adapt to shifting market regimes.
Across three independent runs in out-of-sample rolling backtests on CSI300 and
CSI500, \method achieves mean total returns of 42.93\% and 54.19\%, with mean
Sharpe ratios of 2.46 and 2.86, respectively. It attains the highest mean total
return, Sharpe, and RankIC among the evaluated methods on both universes.
\end{abstract}

\section{Introduction}

Alpha factor mining is the task of discovering predictive signals that rank
assets by their expected future returns
\citep{dingAlphaEvalComprehensiveEfficient2025,tangAlphaAgentLLMDrivenAlpha2025}. 
In formulaic alpha mining, these signals are written as mathematical
expressions over market features, making them easy to inspect, combine, and
deploy inside existing quantitative pipelines
\citep{zhangAutoAlphaEfficientHierarchical2020,yuGeneratingSynergisticFormulaic2023,shiAlphaForgeFrameworkMine2024}. 
This setting poses several general challenges.
First, the search space is combinatorial: even a small library of operators,
features, and lookback windows can generate a huge number of candidate formulas
\citep{zhangAutoAlphaEfficientHierarchical2020,zhuAlphaQCMAlphaDiscovery2025}. 
Second, the supervision is weak and unstable, since financial returns are noisy
and market regimes change over time
\citep{shiAlphaForgeFrameworkMine2024,jiangAlphaR1AlphaScreening2025}. 
Third, a formula that performs well in isolation may become redundant once
added to a larger factor set
\citep{yuGeneratingSynergisticFormulaic2023,tangAlphaAgentLLMDrivenAlpha2025,dingAlphaEvalComprehensiveEfficient2025}. 
A practical alpha-mining framework should therefore use every available
information source, including market-regime descriptions, financial priors,
executable market data, and evaluation feedback, to support
efficient and effective pool-level search.

\begin{table}[t]
\centering
\setlength{\tabcolsep}{4.0pt}
\begin{adjustbox}{max width=\linewidth}
\begin{tabular}{lcccc}
\toprule
Method & Market Text & Financial prior & Real Data & Evaluation Feedback \\
\midrule
GP & -- & \(\triangle\) & \(\triangle\) & -- \\
Neural & -- & -- & \(\checkmark\) & -- \\
LLM-based & \(\checkmark\) & \(\triangle\) & \(\checkmark\) & \(\checkmark\) \\
Alpha-R1 & \(\checkmark\) & \(\checkmark\) & \(\triangle\) & -- \\
\textbf{Ours} & \(\checkmark\) & \(\checkmark\) & \(\checkmark\) & \(\checkmark\) \\
\bottomrule
\end{tabular}
\end{adjustbox}
\caption{Information sources and search capability used by different alpha-mining paradigms. \(\triangle\) denotes partial use: for financial priors, this means heuristic GP design or untuned base-model capability; for real data, this means using market data only through scalar fitness or reward computation.}
\label{tab:info_sources}
\end{table}

Existing methods only use subsets of this information. 
Genetic programming and related symbolic-regression methods produce closed-form
expressions that are easy to inspect and deploy, but they search formula trees
through combinatorial mutation or local search and are typically guided only by
scalar historical fitness, making discovery inefficient and prone to overfitting
\citep{zhangAutoAlphaEfficientHierarchical2020,cranmerInterpretableMachineLearning2023,petersenDeepSymbolicRegression2021,ren2024alphaminingenhancingwarm,guo2026alphaprobealphaminingprincipled}. 
RL and generative systems such as AlphaGen~\citep{yuGeneratingSynergisticFormulaic2023}
and AlphaForge~\citep{shiAlphaForgeFrameworkMine2024} optimize search with
historical market data and downstream rewards, but they do not use language
priors to interpret regimes or explain why a search direction should change
\citep{zhuAlphaQCMAlphaDiscovery2025,xu2024textalpha2discoveringlogicalformulaic,Zhao_2025,zhao2025learningexpertfactorstrajectorylevel}. 
LLM-based frameworks add financial knowledge and natural-language reasoning,
but they still rely heavily on base-model capability and remain closer to
proposal-and-filter pipelines than to practical alpha-mining policies
\citep{liCanLargeLanguage2024,wang2024gptsignalgenerativeaisemiautomated,tangAlphaAgentLLMDrivenAlpha2025,han2026quantaalphaevolutionaryframeworkllmdriven,luo2026alphabench}.
Alpha-R1 \citep{jiangAlphaR1AlphaScreening2025} trains a context-aware
financial reasoning model with GRPO, but it screens a predefined alpha pool
from textual descriptions rather than proposing new formulas or refining them
through executable evaluator feedback.

\begin{figure}[t]
    \centering
    \includegraphics[width=\linewidth]{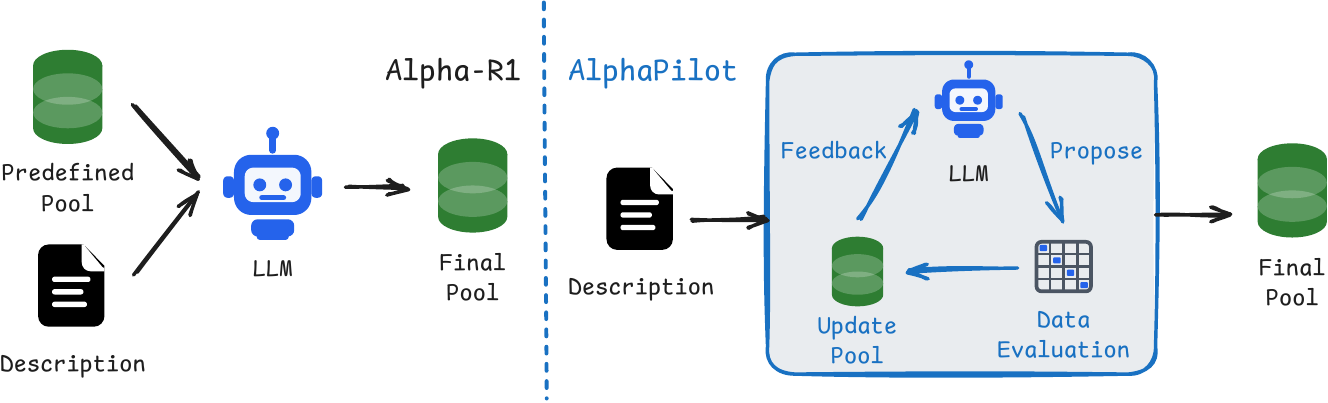}
    \caption{High-level contrast with Alpha-R1. Alpha-R1 screens a predefined
    factor pool from descriptions, while \method closes the loop among proposal,
    data evaluation, feedback, and retained-pool updates.}
    \label{fig:comparison}
\end{figure}

To address this gap, we introduce \method, 
a multi-turn tool-using reinforcement learning framework for adaptive formulaic alpha mining
that integrates the information sources summarized in Table~\ref{tab:info_sources}. 
Like Alpha-R1 \citep{jiangAlphaR1AlphaScreening2025}, 
\method uses an LLM policy conditioned on daily natural-language regime descriptions
and trains it with GRPO.
Unlike one-shot screening, \method gives the agent executable market-data access
and detailed turn-level feedback through a multi-turn tool-use loop.
Figure~\ref{fig:comparison} highlights the contrast with Alpha-R1~\citep{jiangAlphaR1AlphaScreening2025}.
The agent proposes new factors, queries a factor evaluator, updates the factor pool,
observes structured feedback, and refines its proposals through an interpretable search trajectory.  
At inference time, \method mines factors day by day and 
converts the retained pool into a transparent stock-selection signal, 
so the discovered signal set can change with market conditions rather than remain frozen after a single mining stage.

Our main contributions are:
\begin{itemize}
    \item We design a multi-turn tool-using framework for formulaic alpha
    mining. It inherits the ability of prior LLM-based methods to condition
    on market descriptions, while giving the agent executable real-data
    access through a factor-evaluator tool and detailed turn-level feedback
    for newly proposed factors.
    \item We use GRPO to train the agent, improving the financial decision
    ability embedded in a small base model while maintaining inference
    efficiency. This lets the policy learn from multi-turn evaluator
    feedback rather than relying only on the original capability of a large
    general-purpose model.
    \item We build a daily adaptive inference and benchmarking framework for
    alpha-mining methods. It updates factor pools day by day and evaluates
    methods with both factor-level diagnostics and portfolio-level backtests.
\end{itemize}

\section{Related Work}

\subsection{Automated Formulaic Alpha Mining}

Automated formulaic alpha mining has long relied on symbolic search over expression trees. 
Genetic programming and related symbolic regression methods
\citep{zhangAutoAlphaEfficientHierarchical2020,cranmerInterpretableMachineLearning2023,petersenDeepSymbolicRegression2021,ren2024alphaminingenhancingwarm,guo2026alphaprobealphaminingprincipled}
produce closed-form expressions that are easy to deploy and can yield open-ended formulaic novelty through combinatorial search.
However, their search is guided solely by a noisy, non-stationary scalar fitness,
and they cannot incorporate structured evaluator feedback or learn an adaptive search policy across market regimes.

Reinforcement learning and generative-based methods improve this picture by learning search or combination policies 
rather than relying only on unguided symbolic mutation \citep{xu2024textalpha2discoveringlogicalformulaic,Zhao_2025,zhao2025learningexpertfactorstrajectorylevel}. 
AlphaGen \citep{yuGeneratingSynergisticFormulaic2023} and AlphaQCM \citep{zhuAlphaQCMAlphaDiscovery2025} formulate alpha discovery as sequential 
decision making and emphasize the utility of synergistic factor sets. 
AlphaForge \citep{shiAlphaForgeFrameworkMine2024} further introduces learned generation and dynamic combination mechanisms for formulaic factors.
These methods motivate our daily pool-level view of alpha mining, 
but they still rely on weak, noisy, and non-stationary historical scalar rewards,
are less transparent than formula-level language reasoning, and can overfit to backtest-specific feedback. 

Recent financial LLMs, reasoning models, and evaluation benchmarks show that language models can provide financial priors, 
interpretable rationales, and more flexible proposal strategies 
\citep{wu2023bloomberggptlargelanguagemodel,yang2025fingptopensourcefinanciallarge,xie2023pixiulargelanguagemodel,liuFinR1LargeLanguage2025,qianFino1TransferabilityReasoningEnhanced2025,xiaoTradingR1FinancialTrading2025,sunTradeR1BridgingVerifiable2026,luFinToolBenchEvaluatingLLM2026,zhuFinMCPBenchBenchmarkingLLM2026,yangFinVaultBenchmarkingFinancial2026,yuLiveTradeBenchSeekingRealWorld2025,luo2026alphabench}. 
In alpha mining, LLM-based methods have begun to emerge
\citep{wang2024gptsignalgenerativeaisemiautomated,shi2025navigatingalphajunglellmpowered,liu2026cognitivealphaminingllmdriven,kou2025automatestrategyfindingllm,weng2026alphalogicsmarketlogicdrivenmultiagent,han2026quantaalphaevolutionaryframeworkllmdriven}.
Prompt-based systems such as FAMA \citep{liCanLargeLanguage2024} and AlphaAgent \citep{tangAlphaAgentLLMDrivenAlpha2025}
show that LLMs can support autonomous factor generation with natural-language reasoning and
intermediate evaluator signals. However, these systems still rely heavily on base-model capability: stronger models are
slower and more expensive, while insufficiently adapted models can produce hallucinated or weakly grounded proposals.
Alpha-R1 \citep{jiangAlphaR1AlphaScreening2025} is a reinforcement-learning framework tailored to LLM-based alpha mining:
it uses GRPO-trained financial reasoning for context-aware screening. This is an important step toward trainable financial
reasoning, but it still screens a predefined alpha pool rather than designing new factors, relies on textual descriptions
instead of executable data access, and does not provide turn-level evaluator feedback in a multi-turn refinement loop.
Taken together, these limitations motivate a new method that can efficiently use all available information
in the alpha-mining loop, including market descriptions, model priors, real data, and detailed evaluator feedback as summarized in Table~\ref{tab:info_sources}.

\subsection{Agentic RL}
Our formulation is also motivated by recent agentic RL methods that train multi-turn 
tool-using or memory-using LLM policies with downstream outcome rewards, 
including search-augmented reasoning and learned memory construction 
\citep{jinSearchR1TrainingLLMs2025,wangMemaLearningMemory2025,chengAgentR1TrainingPowerful2025}. 
However, alpha mining is different from the QA-style search and memory tasks used in much of this literature. 
Its reward comes from noisy, delayed, and non-stationary market signals rather than discrete answers, 
full backtesting is too slow for every multi-turn trajectory, 
and the agent must track both recent failures and how each new factor changes a retained pool. 
These differences motivate adapting agentic RL frameworks 
to pool-level factor design rather than standard tool-augmented QA.

\section{Problem Setup}

\begin{figure*}[t]
    \centering
    \includegraphics[width=0.80\textwidth]{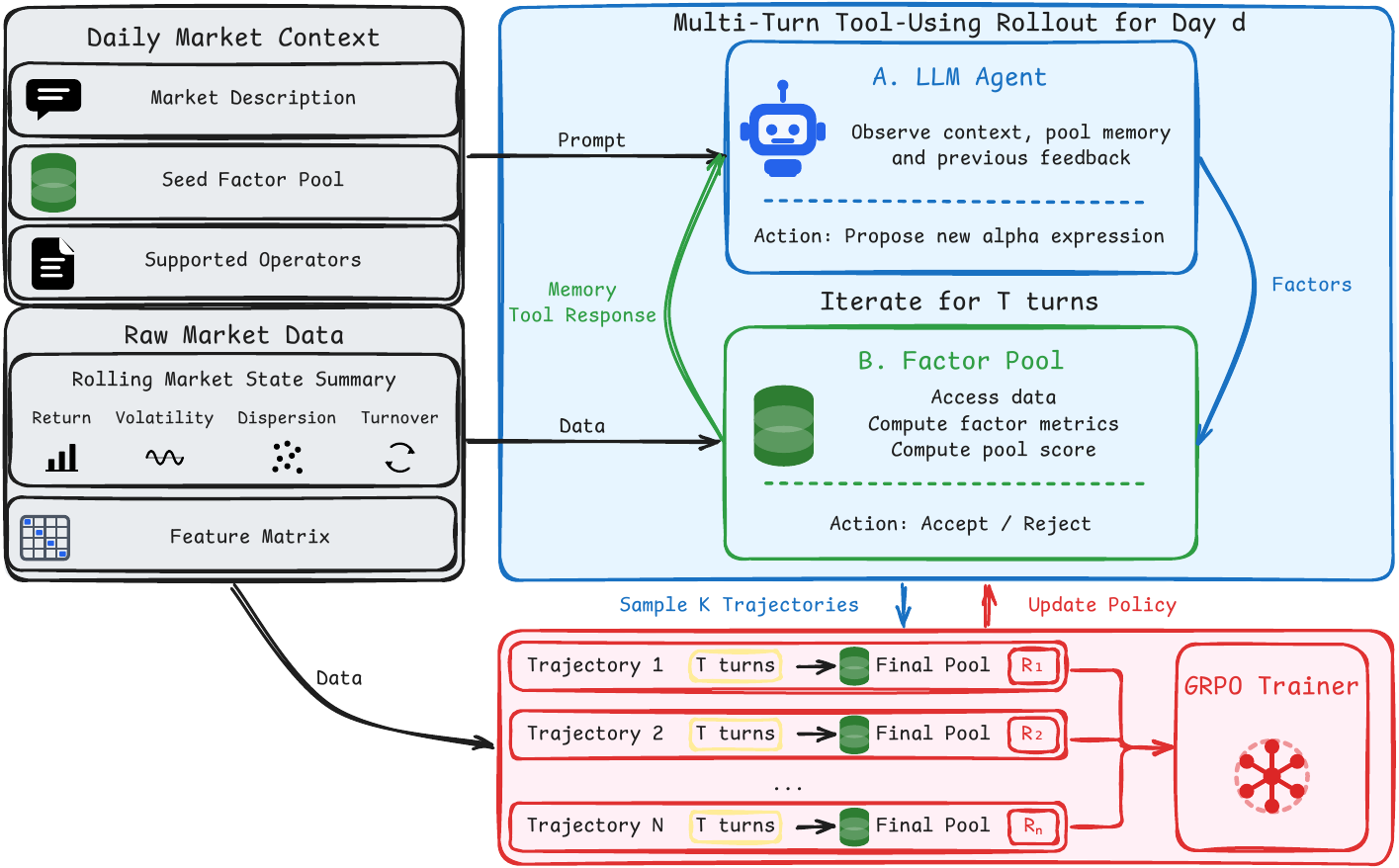}
    \caption{Overview of the \method training pipeline. The agent observes daily market context and pool memory, proposes formulaic alpha expressions through a DSL, receives structured feedback from an executable evaluator, updates a retained factor pool over multiple turns, and is trained with GRPO using trajectory-level rewards computed from the final pool.}
    \label{fig:pipeline}
\end{figure*}

We study formulaic alpha discovery over a stock universe observed on trading days
\(t=1,\ldots,T\). Let \(X_t \in \mathbb{R}^{n_t \times m \times \tau}\)
denote the available market features for the tradable assets on day \(t\), and
let \(y_t \in \mathbb{R}^{n_t}\) denote the next-period return. A formulaic
alpha is a domain-specific language (DSL) expression \(f \in \mathcal{F}\) that maps \(X_t\) to a
cross-sectional score \(v_t=f(X_t)\). The usual single-factor objective asks
whether \(v_t\) ranks assets consistently with \(y_t\), often measured by the
information coefficient and rank information coefficient:
\begin{equation}
\begin{aligned}
\mathrm{IC}_t &= \mathrm{corr}(v_t,y_t), \\
\mathrm{RankIC}_t
&= \mathrm{corr}(\mathrm{rank}(v_t),\mathrm{rank}(y_t)).
\end{aligned}
\end{equation}

In deployment, however, alpha factors are rarely used alone. A general
pool-level objective is to select a set of factors
\(P_t=\{f_{t,1},\ldots,f_{t,K_t}\}\subset\mathcal{F}\) and combination weights
\(w_t\), producing a daily stock-selection signal vector
\begin{equation}
s_t = \sum_{i=1}^{K_t} w_{t,i} f_{t,i}(X_t).
\end{equation}
The optimization task is therefore to construct a factor pool whose aggregated
signal has future predictive utility while satisfying practical constraints
such as validity, coverage, diversity:
\begin{equation}
\begin{aligned}
\max_{P_t,w_t}\;&
\mathcal{J}(P_t,w_t;W_t^{\mathrm{future}})
-\lambda \Omega(P_t,w_t;\mathcal{A}_t) \\
\mathrm{s.t.}\;& f\in\mathcal{F},\quad \forall f\in P_t .
\end{aligned}
\end{equation}
Here \(\mathcal{J}\) can be instantiated by future-window IC/RankIC,
portfolio return, or other downstream utility measures, while
\(\Omega\) denotes a generic regularization term over the candidate pool which
 encourages retained factors to be nonredundant and diverse,
rather than repeatedly proposing variations of the same signal.

\section{AlphaPilot}

\method instantiates the rolling alpha-mining workflow as the multi-turn
tool-using pipeline shown in Figure~\ref{fig:pipeline}. On each decision date,
the system builds a daily market context, lets a language policy interact with
an executable evaluator, and updates a retained factor pool. During offline
training, it computes a reward from the pool's future-window utility. Compared with
prior formula-mining systems, this design is meant to close more of the
information loop: it combines regime-aware language priors, executable
real-market data, detailed turn-level evaluator feedback, and pool-level memory
inside one trainable search process. Appendix Table~\ref{tab:app_method_comparison}
summarizes this high-level contrast.

\subsection{Daily Market Context}
For decision date \(t\), the data pipeline constructs a prompt-level context
\begin{equation}
c_t=(q_t, \mathcal{S}_t, \mathcal{O}),
\end{equation}
where \(q_t\) is the natural-language market-regime description shown to the
agent, \(\mathcal{S}_t\) is the seed or carried-over factor manifest,
and \(\mathcal{O}\) is the supported operator set. 
The seed factor manifest provides candidate factors as starting points, 
while the operator set defines the DSL that the agent is allowed to
use. 

\subsection{Multi-Turn Rollout}
\method casts each daily search as a multi-turn tool-using process. At turn
\(k\), the small LLM policy observes the daily context \(c_t\), compact memory
of recently tried factors, the current retained-pool state, and available tool
feedback from previous turns. It then proposes several DSL factors,
\begin{equation}
a_{t,k}=\{(e_{t,k,j}, g_{t,k,j})\}_{j=1}^{B},
\end{equation}
where \(e_{t,k,j}\) is a formulaic alpha expression and \(g_{t,k,j}\) is a
factor-family label. The system evaluates these factors with executable tools,
updates the retained factor pool, and returns compact feedback to the agent
before the next turn. This loop continues for at most \(H_{\max}\) turns.

\subsection{Retained Factor Pool}

\noindent\textbf{Single-factor evaluation.} The retained-pool module receives new candidate factors through the executable
\texttt{factor\_evaluator}, which is the bridge between language reasoning and
market data. Given a candidate expression, the tool validates the DSL syntax,
executes the factor on the current rolling training window, applies the universe
and tradability filters, and returns single-factor feedback.
The compact feedback shown to the policy includes the validated expression,
factor family, IC, RankIC, stability, and any execution error. 

\noindent\textbf{Factor combination.} Pool-level decisions in \method are made
from the utility of the retained set rather than from isolated factor quality.
To score the current pool, we first combine retained factors into a meta-alpha
by fitting nonnegative weights on the recent training segment:
\begin{equation}
w_t^\star
=\arg\min_{w\ge 0}
\left\|Y_t^{\mathrm{train}}-F(P_{t,k})w\right\|_2^2
+\rho\|w\|_2^2,
\end{equation}
where \(F(P_{t,k})\) is the matrix of retained factor signals. The combined
pool signal scores stocks by
\begin{equation}
s_t(x)=\sum_{f_i\in P_{t,k}}w_{t,i}^\star f_i(x),
\end{equation}
and the scores are converted into cross-sectional ranks before computing the
pool-level diagnostics used by the updater. Appendix~\ref{app:pool_combination_details}
gives the exact standardization and fallback details shared with the seed-factor
baseline.

\noindent\textbf{Pool updater.} After single-factor evaluation, \method applies a lightweight initial filter:
invalid factors, factors with insufficient coverage, and factors with
non-positive single-factor score are discarded. The remaining candidates are
passed to the retained-pool updater, which decides whether each factor should be
added, rejected, or used to replace an existing pool member according to its
marginal effect on the current pool. Let \(\Gamma(P)\) denote the pool score on
the recent training segment:
\begin{equation}
\begin{aligned}
\Gamma(P)
=\;& \frac{1}{2}\mathrm{IC}_{P}
+ \frac{1}{2}\mathrm{RankIC}_{P} \\
&+ \eta_{\mathrm{div}}\mathrm{Div}(P)
+ \eta_{\mathrm{stab}}\mathrm{Stab}(P).
\end{aligned}
\end{equation}
Here \(\mathrm{IC}_{P}\) and \(\mathrm{RankIC}_{P}\) are computed from the
combined pool signal above, while the auxiliary
terms encourage diversity and stability. Following
AlphaEval \citep{dingAlphaEvalComprehensiveEfficient2025}, we compute these
regularization terms in the same way. The full definitions are given in
Appendix~\ref{app:pool_metrics}.

If the pool has free capacity, a surviving
candidate is accepted when it improves \(\Gamma(P)\). If the pool is full, the
candidate must improve \(\Gamma(P)\) after replacing the weakest or most
redundant member. Thus the update
\begin{equation}
P_{t,k+1}=\mathcal{U}(P_{t,k},a_{t,k},\xi_{t,k})
\end{equation}
is driven by marginal contribution to the retained pool rather than by isolated
factor quality alone. 

\noindent\textbf{Feedback.} After all candidates in the current turn are
processed, \method returns three feedback views to the agent. First, a compact
tool response reports the
single-factor metrics and the pool decision for each submitted factor, including
the update action, marginal pool gain, and similarity to the current pool.
Second, a recent-tried memory summarizes prior evaluated candidates with their
expression, family, score, stability, pool-update action, marginal gain, and
similarity to the pool; the current turn's submissions are already covered by
the compact tool responses. This memory allows later turns to avoid repeated
failures. Third, a current-pool memory summarizes the retained pool with its
current pool score and each retained factor's expression, family, score, IC, and
RankIC. These feedback blocks form the history \(h_{t,k}\). They are what make
the search multi-turn rather than proposal-and-filter: the policy can condition
its next proposal on both local evaluator outcomes and the evolving state of the
factor pool.

\subsection{Reward Construction}
At the terminal rollout length \(H\), the final pool \(P_{t,H}\) is passed to a
reward manager. Using the same nonnegative pool-combination rule defined above,
the reward manager forms a final meta-alpha from \(P_{t,H}\), converts it into
cross-sectional ranks, and evaluates the ranked signal on the future window
\(W_t^{\mathrm{future}}\). We then compute
\begin{equation}
\mathrm{PP}_t
= \frac{1}{2}\mathrm{IC}^{\mathrm{future}}_{\mathrm{pool}}
+ \frac{1}{2}\mathrm{RankIC}^{\mathrm{future}}_{\mathrm{pool}}.
\end{equation}
The trajectory reward is
\begin{equation}
R_t = \sigma(\mathrm{PP}_t / \tau)\cdot C(P_{t,H}),
\end{equation}
where \(\sigma\) is the reward mapping used in training, \(\tau\) is a scale
parameter, and \(C(P_{t,H})\) is a pool-completeness term that penalizes
trajectories ending with too few useful retained factors. Appendix~\ref{app:training_params}
gives their closed forms.

\subsection{Policy Optimization}
Finally, \method trains the same tool-using policy with Group Relative Policy
Optimization (GRPO), following recent reasoning-model training practice
\citep{deepseek-aiDeepSeekR1IncentivizingReasoning2025}. For each sampled
decision context \(c_t\), the rollout engine samples a group of \(G\) complete
trajectories \(\{\tau_i\}_{i=1}^{G}\) from the old policy
\(\pi_{\theta_{\mathrm{old}}}\). Each trajectory contains the model's reasoning
tokens, tool calls, evaluator observations, pool updates, and final reward
\(R_i\). GRPO forms a group-relative advantage without training a separate
critic:
\begin{equation}
\hat{A}_i=\frac{R_i-\mathrm{mean}(\{R_j\}_{j=1}^{G})}
{\mathrm{std}(\{R_j\}_{j=1}^{G})+\epsilon}.
\end{equation}
Let \(\rho_{i,u}(\theta)\) be the probability ratio between the current policy
and the old policy for generated token or action \(u\) in trajectory
\(\tau_i\). The clipped surrogate for each token or action is
\begin{equation}
\begin{aligned}
\ell_{i,u}(\theta)
=\min\Big(&\rho_{i,u}(\theta)\hat{A}_i,\\
&\mathrm{clip}(\rho_{i,u}(\theta),1-\epsilon,1+\epsilon)
\hat{A}_i\Big).
\end{aligned}
\end{equation}
The training objective is
\begin{equation}
\begin{aligned}
\mathcal{J}_{\mathrm{GRPO}}(\theta)
=\mathbb{E}\Bigg[
&\frac{1}{G}\sum_{i=1}^{G}
\frac{1}{|\tau_i|}\sum_{u\in\tau_i}\ell_{i,u}(\theta)\\
&-\beta D_{\mathrm{KL}}(\pi_\theta\|\pi_{\mathrm{ref}})
\Bigg].
\end{aligned}
\end{equation}
Optimizing this objective increases the likelihood of proposal, tool-use, and
revision decisions that lead to higher future-window pool rewards relative to
other rollouts from the same market context. This is well matched to alpha
mining: the evaluator metrics at each turn are noisy and local, but the useful
learning signal is delayed until the final factor pool is combined and tested on
future returns.

\FloatBarrier

\begin{table*}[!t]
\centering
\setlength{\tabcolsep}{4pt}
\begin{adjustbox}{max width=\textwidth}
\begin{tabular}{lrrrrr}
\toprule
Method & Total Ret. & Sharpe & Max DD & IC & RankIC \\
\midrule
\method & \textbf{42.93} & \textbf{2.46} & \underline{-9.34} & \textbf{0.0403} & \textbf{0.0595} \\
Qwen3-4B-Instruct-2507 & 0.58 & 0.16 & -16.42 & -0.0165 & -0.0133 \\
Qwen3-30B-A3B-Instruct-2507 & -1.72 & -0.01 & -24.73 & 0.0030 & 0.0146 \\
Qwen/Qwen3-235B-A22B-Instruct-2507 & 6.88 & 0.55 & -16.32 & 0.0093 & 0.0226 \\
DeepSeek-R1 direct & 22.79 & 1.21 & -16.20 & 0.0114 & -0.0033 \\
Seed factors + LR & 34.51 & 1.82 & -18.65 & 0.0366 & 0.0361 \\
AlphaGen & 25.17 & 2.26 & \textbf{-8.24} & -0.0009 & 0.0109 \\
AlphaForge & 2.18 & 0.07 & -12.15 & -0.0086 & 0.0024 \\
\bottomrule
\end{tabular}
\end{adjustbox}
\caption{CSI300 out-of-sample results. Returns and drawdowns are percentages. Bold marks the best value; underline marks \method when second best.}
\label{tab:csi300_main}
\end{table*}

\begin{table*}[!t]
\centering
\setlength{\tabcolsep}{4pt}
\begin{adjustbox}{max width=\textwidth}
\begin{tabular}{lrrrrr}
\toprule
Method & Total Ret. & Sharpe & Max DD & IC & RankIC \\
\midrule
\method & \textbf{54.19} & \textbf{2.86} & -14.34 & \textbf{0.0259} & \textbf{0.0624} \\
Qwen3-4B-Instruct-2507 & -4.83 & -0.30 & -21.60 & -0.0207 & 0.0090 \\
Qwen3-30B-A3B-Instruct-2507 & 10.42 & 0.89 & -12.79 & 0.0064 & 0.0409 \\
Qwen/Qwen3-235B-A22B-Instruct-2507 & 7.10 & 0.54 & -16.17 & -0.0056 & 0.0064 \\
DeepSeek-R1 direct & 14.07 & 0.84 & -18.60 & 0.0042 & -0.0211 \\
Seed factors + LR & 18.56 & 1.07 & -22.67 & 0.0201 & 0.0189 \\
AlphaGen & 48.85 & 2.82 & -13.45 & 0.0060 & 0.0042 \\
AlphaForge & 36.32 & 1.97 & \textbf{-11.11} & -0.0091 & 0.0086 \\
\bottomrule
\end{tabular}
\end{adjustbox}
\caption{CSI500 out-of-sample results. Returns and drawdowns are percentages. Bold marks the best value; underline marks \method when second best.}
\label{tab:csi500_main}
\end{table*}

\section{Experiments}

\subsection{Experimental Setup}

\paragraph{Market data and examples.}
We use daily Chinese A-share panels for CSI300 and CSI500. Each decision
example contains a 3-year rolling history window, natural-language regime
descriptions, and a seed factor library. The 21- and 63-day future windows
serve offline evaluation only and are unavailable to the inference policy.
We split at \texttt{2024-01-01} for validation and
\texttt{2025-07-01} for testing, sample train/validation examples every 5
trading days, and run daily test inference. Appendix
Table~\ref{tab:app_data_splits} gives the exact date ranges, example counts,
sampling schedules, and split objectives.

\paragraph{Compared methods.}
We compare \method with deterministic LLM-agent baselines using
\texttt{Qwen/Qwen3-4B-Instruct-2507},
\texttt{Qwen/Qwen3-30B-A3B-Instruct-2507}, and
\texttt{Qwen/Qwen3-235B-A22B-Instruct-2507}, all decoded with temperature 0.
Following the benchmarking results in AlphaEval
\citep{dingAlphaEvalComprehensiveEfficient2025}, we use
AlphaGen~\citep{yuGeneratingSynergisticFormulaic2023} as the RL-based baseline
and AlphaForge~\citep{shiAlphaForgeFrameworkMine2024} as the generative-based
baseline.
For CSI300, we also include a controlled GRPO single-turn baseline with the
same Qwen3-4B backbone, GRPO algorithm, data splits, and terminal pool-level
reward as \method. It generates the final factor pool in one response, removing
iterative tool-based refinement while retaining RL training.
We also include seed-factor linear regression and a DeepSeek-R1
direct-propose reference. We do not report a reproduced AlphaAgent baseline
because the available open-source release does not provide enough detail for
faithful replication in our setting. Appendix~\ref{app:baselines} describes the baseline
pipelines, seed factor library, and reproduction notes.

\paragraph{Training and inference.}
\method is trained with GRPO in VERL~\cite{sheng2024hybridflow} using
\texttt{Qwen/Qwen3-4B-Instruct-2507}. The main setup samples eight rollouts per
prompt, allows up to 40 assistant turns, and uses a sigmoid-mapped
future-window IC/RankIC reward with a minimum-positive-factor completeness
penalty. At inference time, we preserve each baseline's original
factor-generation design as much as possible and evaluate all methods under a
common day-by-day portfolio protocol. \method uses recent-window
NNLS/ridge-style factor weights; AlphaGen and AlphaForge retain their native
weights (Appendix~\ref{app:baselines}). All signals are traded as long-only
top-\(K\) portfolios with 10 bps transaction costs.

\paragraph{Metrics.}
The main tables report total return, Sharpe, maximum drawdown, IC, and RankIC.
Returns and drawdowns are percentages; IC and RankIC are daily cross-sectional
metrics computed from the traded signal and next-day returns. We report
three-run means for \method, GRPO single-turn, AlphaGen, and AlphaForge, and
single-run results for other methods; annualized returns and standard deviations
are in Appendix Table~\ref{tab:app_main_results_detailed}. For the
benchmark-relative analysis, we construct an equal-weight benchmark from the
same tradable universe over the same test period and compute daily excess
returns as portfolio returns minus benchmark returns. Cumulative excess return
is the difference between cumulative portfolio and benchmark returns.

\begin{figure*}[!t]
    \centering
    \includegraphics[width=0.98\textwidth]{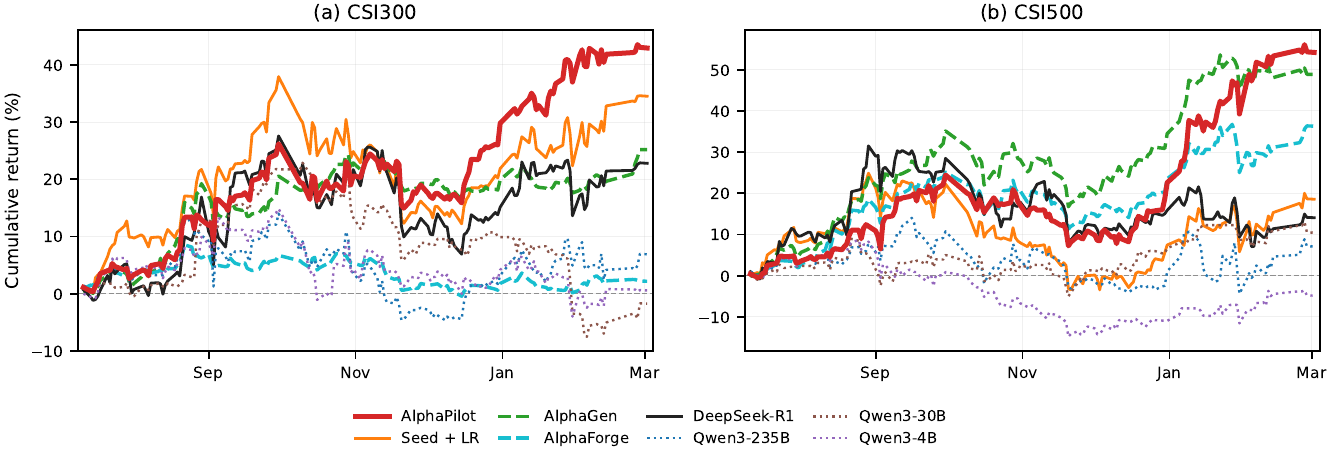}
\caption{Out-of-sample cumulative return curves on CSI300 and CSI500 from
    July 2025 to March 2026 under the same portfolio protocol. Stochastic
    curves for \method, AlphaGen, and AlphaForge show three-run means. The
    panels use different y-axis scales.}
    \label{fig:equity_curves}
\end{figure*}

\begin{table*}[tp]
\centering
\setlength{\tabcolsep}{3pt}
\begin{adjustbox}{max width=0.9\textwidth}
\begin{tabular}{lrrrrrr}
\toprule
Method / Variant & Avg. Turns & Total Ret. & Sharpe & Max DD & IC & RankIC \\
\midrule
\multicolumn{7}{l}{\textit{Controlled multi-turn comparison (three-run means)}} \\
\method & \textemdash & \textbf{42.93} & \textbf{2.46} & \textbf{-9.34} & \underline{0.0403} & \textbf{0.0595} \\
GRPO single-turn & \textemdash & 35.62 & 2.02 & -16.32 & \textbf{0.0407} & 0.0496 \\
\midrule
\multicolumn{7}{l}{\textit{Matched single-run design ablations}} \\
Full method & \textbf{5.24} & \textbf{48.64} & \textbf{2.74} & \underline{-10.00} & 0.0361 & 0.0501 \\
Full + reward auxiliary & 5.61 & 44.49 & 2.50 & -10.00 & 0.0408 & 0.0594 \\
No pool auxiliary score & 5.67 & 34.39 & 2.06 & \textbf{-8.53} & \textbf{0.0419} & \textbf{0.0634} \\
No memory feedback & 15.75 & 26.40 & 1.70 & -10.99 & 0.0293 & 0.0473 \\
\bottomrule
\end{tabular}
\end{adjustbox}
\caption{CSI300 controlled comparisons and design ablations. Returns and
drawdowns are percentages. Avg. Turns, reported for the design ablations, is the
average number of assistant turns per daily rollout. Bold marks the best value
within each comparison group (lower for Avg. Turns; higher otherwise);
underline marks \method or the full method when it is second best. Detailed
results are reported in Appendix Table~\ref{tab:app_csi300_ablation_detailed}.}
\label{tab:csi300_ablation}
\end{table*}

\subsection{Main Comparison Against Baselines}

\paragraph{CSI300.}
Table~\ref{tab:csi300_main} shows that \method is the strongest method on
CSI300 by mean total return, Sharpe, and RankIC. It
achieves 42.93\% total return, 2.46 Sharpe, 0.0403 IC, and 0.0595 RankIC. The
seed-factor linear-regression baseline remains
substantially below \method in portfolio return, Sharpe, and signal quality.
The deterministic LLM baselines are much weaker, suggesting that prompting alone
does not reliably turn a base model into a useful daily alpha-mining policy.

\paragraph{CSI500.}
Table~\ref{tab:csi500_main} gives the corresponding CSI500 results. \method
obtains the highest mean total return, Sharpe, IC, and RankIC among the reported
main-table methods, reaching 54.19\% total return and 2.86 Sharpe. AlphaGen and
AlphaForge show substantial
variation across runs on this universe, while \method produces the strongest
average signal quality in the table.

\paragraph{Benchmark-relative performance.}
Against an equal-weight benchmark from the same tradable universe, \method
achieves 21.49\% cumulative excess return with a 1.65 excess Sharpe on CSI300,
and 18.22\% cumulative excess return with a 1.36 excess Sharpe on CSI500.
Appendix Table~\ref{tab:app_benchmark_relative} reports the complete results.

\paragraph{Equity curves.}
Figure~\ref{fig:equity_curves} plots cumulative return over the test period.
On CSI300, its mean trajectory recovers from a mid-period decline and rises
strongly in the final months. On CSI500, it closes the gap with AlphaGen late in
the test period. It therefore records the highest final value on both CSI300
and CSI500.

\subsection{CSI300 Ablation Studies}

The controlled-comparison block of Table~\ref{tab:csi300_ablation}
isolates the contribution of iterative refinement beyond RL training. Compared
with GRPO single-turn, the three-run \method result improves mean total return
by 7.31 percentage points, Sharpe by 0.43, maximum drawdown by 6.98 percentage
points, and RankIC by 0.0099, while achieving comparable IC (0.0403 versus
0.0407). Differences are computed from the unrounded run means. Both settings
use the same backbone, GRPO algorithm, data splits, and terminal pool-level
reward.

The design-ablation block of Table~\ref{tab:csi300_ablation} compares the full method with three
single-run design ablations on CSI300. In this comparison, the full method
obtains the strongest portfolio return and Sharpe among the reported variants.

Removing memory feedback lowers total return from 48.64\% to
26.40\% and increases average rollout length from 5.24 to 15.75 turns. This
suggests that compact tried-factor and retained-pool feedback is not only useful
for selecting better factors, but also for steering the agent away from longer,
less productive search trajectories.

Adding a separate reward-side auxiliary term gives comparable but slightly
weaker portfolio results: 44.49\% total return and 2.50 Sharpe, compared with
48.64\% and 2.74 for the full setting. We therefore keep auxiliary terms in the
retained-pool score, where they directly affect pool selection, but omit the
extra reward-side auxiliary term in the main configuration for efficiency.

The no-pool-auxiliary variant has the best IC and RankIC in Table~\ref{tab:csi300_ablation},
but weaker total return and Sharpe than the full method. This gap shows that the
training proxy need not align with realized portfolio return after weighting,
selection, and transaction costs. We therefore prioritize pool-level utility
over IC or RankIC alone.

\subsection{Qualitative Analysis}

\begin{table}[!htbp]
\centering
\setlength{\tabcolsep}{2pt}
\begin{adjustbox}{max width=\linewidth}
\begin{tabular}{rrlrr}
\toprule
Add \# & Turn & Added factor & Score & Pool score \\
\midrule
1 & 1 & \texttt{-rank(ts\_delta(close, 1))} & 0.380 & 0.0432 \\
2 & 1 & \texttt{-rank(close / delay(close, 5) - 1)} & 0.171 & 0.0661 \\
3 & 2 & \texttt{-rank(ts\_delta(close, 3))} & 0.190 & 0.0665 \\
4 & 2 & \texttt{-rank(ts\_delta(close, 7))} & 0.0329 & 0.0674 \\
5 & 2 & \texttt{-rank(ts\_delta(close, 15))} & 0.0494 & 0.0674 \\
6 & 2 & \texttt{-rank(ts\_corr(close, volume, 10))} & 0.355 & 0.0727 \\
7 & 2 & \texttt{-rank(ts\_corr(close, volume, 5))} & 0.136 & 0.0727 \\
8 & 2 & \texttt{-rank(vwap\_spread)} & 0.637 & 0.0813 \\
9 & 4 & \texttt{-rank(close / volume)} & 0.473 & 0.0844 \\
10 & 4 & \texttt{-rank(ts\_delta(volume, 5))} & 0.336 & 0.0885 \\
11 & 4 & \texttt{-rank(ts\_delta(volume, 10))} & 0.386 & 0.0885 \\
12 & 4 & \texttt{-rank(ts\_delta(close, 30))} & 0.0834 & 0.0917 \\
13 & 4 & \texttt{-rank(ts\_corr(close, volume, 20))} & 0.270 & 0.0931 \\
14 & 4 & \texttt{-rank(ts\_corr(close, volume, 30))} & 0.153 & 0.0938 \\
\bottomrule
\end{tabular}
\end{adjustbox}
\caption{Factors added to the retained pool during the CSI300 \method rollout on the first formal test trading day. The final column records the cumulative retained-pool score after each accepted addition; rejected tool calls are omitted.}
\label{tab:case_study_process}
\end{table}

Beyond aggregate metrics, we inspect the CSI300 \method rollout on the first
formal test trading day, 2025-07-11. The market description suggests a
supportive uptrend, so momentum and trend-following factors are plausible first
hypotheses. However, the evaluator assigns weak or negative RankIC to the pure
momentum/trend proposals, after which the agent pivots toward reversal and
liquidity-sensitive factors. Table~\ref{tab:case_study_process} lists the
factors that are actually added to the retained pool.

The trajectory highlights four behaviors. First, tool feedback corrects the
initial regime prior: although the market description suggests momentum, the pool
is ultimately dominated by reversal-style factors. Second, the strongest
retained factors are interpretable liquidity-driven reversal signals, including
\texttt{-rank(vwap\_spread)}, \texttt{-rank(close / volume)}, and
\texttt{-rank(ts\_delta(volume, 10))}. Third, several volatility-adjusted
variants receive high standalone IC/RankIC but are rejected because their
marginal pool gain is negative (see Table~\ref{tab:app_case_submissions} in
Appendix~\ref{app:case_submission_trace}). The final pool therefore reflects
ensemble utility rather than single-factor strength alone, reaching a pool score
of 0.0938 with 14 retained factors in 4 turns. Fourth, only 4 of these 14
accepted factors come directly from the seed library, while the other 10 are
newly proposed mutations, showing that the agent actively expands the factor set
rather than merely selecting from seeds.

\section{Conclusion}

We presented \method, a multi-turn tool-using RL framework for formulaic alpha
mining that combines market descriptions, LLM financial priors, executable
market data, and turn-level evaluator feedback while optimizing a retained
factor pool. Across CSI300 and CSI500 rolling backtests, it achieves the highest
mean total return, Sharpe, and RankIC among the evaluated methods on both
universes, offering a scalable path for agentic alpha-mining policy training
and inference.

\section*{Limitations}

Our experiments are limited by the scale of the models and benchmarks we can
afford to run. Because GRPO training with multi-turn tool use is
computationally expensive, we train \method with a small open-weight LLM
instead of scaling the same procedure to larger policies. Similarly, due to API
and financial constraints, we do not benchmark top proprietary LLMs in the
full multi-turn tool-using setting. Evaluation covers one contiguous period
(July 2025--March 2026) in two Chinese equity universes; broader historical
and cross-market generalization remains untested.

Our reward uses short-term future-window IC/RankIC proxy metrics rather than
executing a full backtest inside every training trajectory. This proxy keeps training
tractable, but it may miss deployment objectives such as drawdown control,
capacity, liquidity, transaction-cost sensitivity, and risk exposure. An
important direction is to design efficient reward pipelines that incorporate
richer backtesting signals during alpha mining without making RL training
prohibitively slow.

Finally, our goal is to study multi-turn tool-using RL training
 for LLM-based alpha mining, not to optimize a complete trading
strategy. We therefore use a simple and transparent evaluation
protocol---recent-window factor combination, a long-only top-\(K\) portfolio,
daily rebalancing, and stylized transaction costs---to test whether the
discovered factor pools are useful under common conditions. Designing stronger
portfolio construction rules and trading strategies is left for future work.

\section*{Ethical Considerations}

This work studies LLM-assisted alpha factor discovery for research purposes.
The proposed framework is intended for offline factor exploration and analysis,
rather than as investment advice or an autonomous trading system.
Any real-world deployment would require additional validation, risk control,
compliance review, transaction-cost modeling, market-impact analysis, and live monitoring.

Our experiments use historical market data and do not involve personal or
user-level private information.
The future-window evaluation is used only to compute offline training rewards
and is not available to the agent during deployment.

\bibliography{ref/alpha,ref/symbolic,ref/emnlp2026_alpha_agent,ref/emnlp2026_alpha_finance,ref/addtional_alpha,ref/addtional_alpha_2026}

\clearpage
\appendix
\section*{Appendix Contents}
This appendix provides implementation details, efficiency analysis, a detailed
case-study trace, and prompt and feedback templates that support the main paper.

\noindent\textbf{Appendix~\ref{app:implementation}. Implementation Details.}
\newline
\hspace*{1em}\textbf{Appendix~\ref{app:pool_metrics}.} Retained-pool diversity and stability metrics.
\newline
\hspace*{1em}\textbf{Appendix~\ref{app:pool_combination_details}.} Shared factor-combination and signal-construction rule.
\newline
\hspace*{1em}\textbf{Appendix~\ref{app:dataset_construction}.} Dataset setup and data processing pipeline.
\newline
\hspace*{1em}\textbf{Appendix~\ref{app:supported_operators}.} Supported DSL operators.
\newline
\hspace*{1em}\textbf{Appendix~\ref{app:training_params}.} Training and inference parameters.
\newline
\hspace*{1em}\textbf{Appendix~\ref{app:baselines}.} Baseline descriptions.
\newline
\hspace*{1em}\textbf{Appendix~\ref{app:detailed_results}.} Detailed multi-run results.
\newline
\hspace*{1em}\textbf{Appendix~\ref{app:ablation_design}.} Ablation design.

\medskip
\noindent\textbf{Appendix~\ref{app:efficiency_training}. Efficiency, Pool Behavior, and Training Dynamics.}
\newline
\hspace*{1em}\textbf{Appendix~\ref{app:inference_efficiency}.} Inference-time efficiency.
\newline
\hspace*{1em}\textbf{Appendix~\ref{app:training_dynamics}.} GRPO training dynamics.
\newline
\hspace*{1em}\textbf{Appendix~\ref{app:avg_pool_size}.} Average daily pool size.
\newline
\hspace*{1em}\textbf{Appendix~\ref{app:seed_pool_similarity}.} Seed-pool similarity analysis.

\medskip
\noindent\textbf{Appendix~\ref{app:case_submission_trace}. Case-Study Submission Trace.}
\newline
\hspace*{1em}\textbf{Appendix~\ref{app:complete_submission_trace}.} Complete evaluated factor submissions.

\medskip
\noindent\textbf{Appendix~\ref{app:prompts}. Prompt and Feedback Templates.}
\newline
\hspace*{1em}\textbf{Appendix~\ref{app:multi_turn_prompt}.} Multi-turn tool-using prompt.
\newline
\hspace*{1em}\textbf{Appendix~\ref{app:example_market_description}.} Example market description used by the agent.
\newline
\hspace*{1em}\textbf{Appendix~\ref{app:market_verbalization_prompt}.} Market-state verbalization prompt.
\newline
\hspace*{1em}\textbf{Appendix~\ref{app:tool_feedback_prompt}.} Tool feedback template.

\FloatBarrier
\section{Implementation Details}
\label{app:implementation}

This section gives reproducibility details that are too implementation-heavy for
the main paper. We first define the pool diagnostics used by the retained-pool
updater, then describe the shared factor-combination rule, data construction,
training/inference settings, baseline protocols, and the seed-factor library
used as a common starting point.

\begin{table*}[t]
\centering
\setlength{\tabcolsep}{5pt}
\begin{adjustbox}{width=\textwidth,center}
\begin{tabular}{lcccccccc}
\toprule
Method & Model & Regime aware & Refinement & Pool & Training & Daily inference & Efficiency & New factor design \\
\midrule
AlphaGen & Neural & No & No & Yes & Yes & No & High & Yes \\
AlphaForge & Neural & No & No & Yes & Yes & Weak & Medium & Yes \\
AlphaAgent & LLM & Yes & Yes & No & No & Weak & Low & Yes \\
Alpha-R1 & LLM & Yes & No & Yes & Yes & Yes & Medium & No \\
\method & \textbf{LLM} & \textbf{Yes} & \textbf{Yes} & \textbf{Yes} & \textbf{Yes} & \textbf{Yes} & \textbf{Medium} & \textbf{Yes} \\
\bottomrule
\end{tabular}
\end{adjustbox}
\caption{High-level comparison of representative alpha-mining frameworks. \method is the only method in this comparison that combines regime awareness, iterative refinement, retained-pool optimization, task-specific training, chronological daily inference, and new-factor design in one framework. ``Refinement'' indicates whether the method supports iterative revision from intermediate feedback inside one daily search trajectory. ``Daily inference'' indicates whether the method naturally supports chronological day-by-day deployment rather than one-shot mining only. ``Efficiency'' is a qualitative summary of practical deployment cost, especially repeated day-by-day LLM inference.}
\label{tab:app_method_comparison}
\end{table*}

\subsection{Pool Diversity and Stability Metrics}
\label{app:pool_metrics}

For consistency with prior alpha-mining evaluation, we compute retained-pool
diversity and stability in the same style as AlphaEval
\citep{dingAlphaEvalComprehensiveEfficient2025}. Let \(R_i\) be the time-by-stock
rank-signal matrix of factor \(f_i\) on the recent training segment, and let
\(z_i=\mathrm{vec}(R_i)\). We define diversity as the normalized entropy of the
covariance spectrum:
\begin{equation}
\begin{aligned}
\mathrm{Div}(P)
&= -\frac{1}{\log K}
\sum_{\ell=1}^{K} p_\ell \log p_\ell,\\
p_\ell
&=\lambda_\ell\Big/\sum_{r=1}^{K}\lambda_r,
\end{aligned}
\end{equation}
where \(\{\lambda_\ell\}_{\ell=1}^{K}\) are the eigenvalues of
\(\mathrm{Cov}(\{z_i\}_{i=1}^{K})\). This score is small when retained factors
collapse to redundant rank patterns and larger when their signals span multiple
directions.

For stability, let \(r_{i,d}\) denote the cross-sectional rank vector of factor
\(f_i\) on day \(d\). The factor-level stability is the average day-to-day rank
correlation,
\begin{equation}
\mathrm{Stab}(f_i)
= \frac{1}{|D|-1}\sum_{d=2}^{|D|}
\mathrm{corr}(r_{i,d}, r_{i,d-1}),
\end{equation}
and the pool stability is a weighted average over retained factors:
\begin{equation}
\mathrm{Stab}(P)=
\sum_{i=1}^{K} \bar{w}_i\,\mathrm{Stab}(f_i),
\quad
\bar{w}_i=\frac{w_i}{\sum_{j=1}^{K}w_j}.
\end{equation}

\subsection{Shared Factor Combination and Signal Construction}
\label{app:pool_combination_details}

The main paper omits several implementation details of how a retained factor
set is turned into a tradable stock-selection signal. This procedure is shared
by \method and the seed-factor linear-regression baseline, so we describe it
once here. Given a pool of validated factors and a rolling training window, we
first evaluate each factor over the recent 126 trading days of that window. For
each day, the factor cross-section is converted to a rank-percentile signal and
shifted to the range \([-0.5, 0.5]\). These rank signals are the inputs used
both for recent-window weight fitting and for same-day signal construction at
inference time.

We then fit nonnegative combination weights on the recent window using the same
NNLS/ridge-style objective described in the method section. If all retained
factors can be evaluated on a common recent window, the fitted weights are
normalized to sum to one and used as the pool weights. If the fit is
ill-conditioned, too few valid samples are available, or some factors fail
window-level evaluation, we fall back to positive score-based weights derived
from the factor scores stored in the pool. If even these scores are unusable,
we use equal weights over the surviving factors.

On each decision day, the current cross-section of every usable factor is again
converted to a rank-percentile signal in \([-0.5, 0.5]\), and the final
meta-alpha is the weighted sum of these rank signals. The resulting stock-level
signal is then used both for pool-level diagnostics such as IC and RankIC and
for the long-only top-\(K\) portfolio backtest. This shared construction is why
the main paper can compare \method and the seed-factor linear-regression
baseline under the same portfolio interface without conflating differences in
factor search with differences in downstream signal aggregation.

\subsection{Dataset Construction and Processing Pipeline}
\label{app:dataset_construction}
\label{app:data_processing}

Table~\ref{tab:app_dataset_setup} summarizes the data setup used in our
experiments, while Table~\ref{tab:app_data_splits} reports the chronological
split. The same preprocessing pipeline is used for both universes: it starts
from daily stock panels, applies universe and tradability filters, builds
stock-level features, aggregates market-state statistics, and converts them
into natural-language regime descriptions used in the agent prompt.

\begin{table}[!htbp]
\centering
\begin{adjustbox}{max width=\linewidth}
\begin{tabular}{ll}
\toprule
Item & Setting \\
\midrule
Universes & CSI300, CSI500 \\
Frequency & Daily \\
Training window per example & 3 years \\
Future evaluation windows & 21 and 63 trading days \\
Regime summary horizons & 10, 63, and 252 trading days \\
Validation split & 2024-01-01 \\
Test split & 2025-07-01 \\
Purge gap & 10 trading days \\
Warmup before test & 10 trading days \\
Train/validation sampling & Every 5 trading days \\
Test inference & Every trading day \\
\bottomrule
\end{tabular}
\end{adjustbox}
\caption{Dataset and rolling-example construction.}
\label{tab:app_dataset_setup}
\end{table}

\begin{table*}[t]
\centering
\setlength{\tabcolsep}{4pt}
\begin{adjustbox}{width=\textwidth,center}
\begin{tabular}{lp{0.29\textwidth}p{0.20\textwidth}p{0.34\textwidth}}
\toprule
Split & Decision-date range & Count / sampling & Objective \\
\midrule
Training &
2011-06-08--2023-09-20 &
599 per universe &
Optimize the policy with GRPO using future-window retained-pool rewards. \\
Validation &
2024-01-12--2025-03-24 &
58 per universe &
Checkpoint/model selection and hyperparameter sanity checking without gradient
updates. \\
Testing &
2025-07-11--2026-03-02 &
153 per universe &
Chronological out-of-sample inference and portfolio backtesting. \\
\bottomrule
\end{tabular}
\end{adjustbox}
\caption{Detailed dataset split construction used for CSI300 and CSI500. Each
decision example contains a 3-year rolling history window and future
evaluation windows of 21 and 63 trading days.}
\label{tab:app_data_splits}
\end{table*}

The raw panel includes price, volume, amount, VWAP, turnover, valuation, and
status fields. The feature store includes returns, rolling volatility, moving
averages, volume statistics, relative price-volume features, and sector
metadata when available. A stock is tradable only when it is in the current
universe, has passed the IPO seasoning filter, is not suspended, and is not
marked ST.

\paragraph{Processing pipeline.}
We run the same preprocessing pipeline separately for CSI300 and CSI500 by
setting the active universe flag. The pipeline starts from Qlib daily panels
over 2008--2026. For each target universe, it collects all stocks that ever
enter the universe, extracts OHLCV, amount, VWAP, turnover, adjustment factor,
PE/PB valuation fields, and ST status, and constructs a time-varying
\texttt{is\_\{universe\}\_member\_t} membership indicator from index membership
intervals.

The raw panel is cleaned before feature construction. Rows with all OHLCV fields
missing are removed, and suspension days are identified by zero or missing
volume. Prices and adjustment factors are forward-filled by stock, missing VWAP
is filled with the cleaned close, volume/amount/turnover are filled with zero,
and low-frequency valuation/ST fields are forward-filled. A stock is treated as
tradable only when it has at least 60 historical trading days, is not suspended,
and is not ST.

From the cleaned panel, we build the daily feature store used by the DSL and
market-state summaries. The stock-level features include \texttt{ret\_1d},
\texttt{ret\_5d}, \texttt{ret\_20d}, \texttt{vol\_20d}, \texttt{ma20},
\texttt{ma50}, \texttt{vol\_mean\_20d}, \texttt{vol\_std\_20d},
\texttt{turn\_mean\_20d}, \texttt{price\_to\_ma20},
\texttt{price\_to\_ma50}, \texttt{volume\_ratio}, and \texttt{vwap\_spread}.
Sector labels are attached when available and translated to English for prompt
readability.

Daily market-state metrics are aggregated only over stocks that are both
tradable and current universe members. These summaries include equal-weighted
market returns, 5/10/20-day compound returns, return dispersion, 5/20-day market
volatility, the fraction of rising stocks, fractions above 20/50-day moving
averages, 5-day changes in volume and turnover, and strongest/weakest sector
rotation statistics.

Finally, we convert the market metrics into rolling decision examples. Each
example contains 10-day short-term, 63-day medium-term, and 252-day long-term
market summaries, a 756-trading-day historical training window, and 21/63-day
future evaluation windows. Training and validation candidates are generated every
5 trading days; test examples are generated daily and include a 10-trading-day
warmup before the evaluation period. Structured market states are converted into
four-field English regime descriptions using Llama-3.3-70B-Instruct
with temperature 0. The resulting examples are split chronologically with
a 10-day purge gap and converted into training instances containing the prompt,
seed factor library, supported operators, reward metadata, train/test windows,
and \texttt{factor\_evaluator} tool context.

\subsection{Supported DSL Operators}
\label{app:supported_operators}

The evaluator enforces a fixed expression grammar so that every proposed alpha
can be parsed and executed consistently. Table~\ref{tab:app_operator_block}
lists the complete operator set exposed to the agent.

\begin{table}[H]
\centering
\small
\setlength{\tabcolsep}{4pt}
\begin{adjustbox}{max width=\linewidth}
\begin{tabular}{lp{0.72\linewidth}}
\toprule
Category & Operators \\
\midrule
Cross-sectional & \texttt{rank} \\
Time-series &
\texttt{ts\_mean}, \texttt{ts\_sum}, \texttt{ts\_std}, \texttt{ts\_var},
\texttt{ts\_delta}, \texttt{delay}, \texttt{ts\_cov}, \texttt{ts\_corr},
\texttt{ts\_rank}, \texttt{ts\_zscore}, \texttt{ts\_min}, \texttt{ts\_max},
\texttt{ts\_median}, \texttt{ts\_mad} \\
Smoothing & \texttt{decay\_linear}, \texttt{ema} \\
Transforms & \texttt{greater}, \texttt{less}, \texttt{abs}, \texttt{log}, \texttt{sign} \\
Arithmetic & \texttt{+}, \texttt{-}, \texttt{*}, \texttt{/} \\
\bottomrule
\end{tabular}
\end{adjustbox}
\caption{Complete DSL operator set available to the agent.}
\label{tab:app_operator_block}
\end{table}

\subsection{Training and Inference Parameters}
\label{app:training_params}

Table~\ref{tab:app_training_setup} records the main training and inference configuration used
by the reported \method runs rather than the script defaults. We use VERL with
multi-turn tool use and the SGLang rollout backend.

For the main configuration without reward-side auxiliary terms, the mapping
in the trajectory reward is
\begin{equation}
\sigma(z)=\operatorname{clip}\!\left(\frac{2}{1+\exp(-z)}-1,0,1\right).
\end{equation}
The completeness multiplier is
\begin{equation}
C(P)=0.5+0.5\min\!\left(\frac{n_{+}(P)}{n_{\min}},1\right),
\end{equation}
where \(n_{+}(P)\) counts retained factors with a positive raw factor score.
We use \(\tau=0.02\) and \(n_{\min}=10\), so
\(R(P)=\sigma(\mathrm{PP}(P)/\tau)C(P)\).
Nonpositive predictive scores map to zero reward; the completeness multiplier
ranges from 0.5 to 1. Empty or invalid pools receive zero reward.

\begin{table}[t]
\centering
\begin{adjustbox}{max width=\linewidth}
\begin{tabular}{ll}
\toprule
Item & Setting \\
\midrule
Base policy & \texttt{Qwen/Qwen3-4B-Instruct-2507} \\
RL algorithm & GRPO \\
Rollouts per prompt & 8 \\
Maximum assistant turns & 40 \\
Maximum parallel tool calls & 8 \\
Maximum prompt length & 4096 tokens \\
Maximum response length & 10240 tokens \\
Inference temperature & 0.0 \\
Training batch size & 32 \\
PPO minibatch size & 32 \\
Actor learning rate & \(1\times 10^{-6}\) \\
KL loss coefficient & 0.001 \\
Reward mapping & Sigmoid \\
Sigmoid scale & 0.02 \\
Minimum positive factors & 10 \\
Pool auxiliary in score & True \\
Reward auxiliary in reward & False \\
Memory feedback & True \\
Training epochs & 5 \\
Main retained-pool limit & 20 \\
Pool scoring window & Recent 126 train days \\
Portfolio top-\(K\) & 20 stocks \\
Transaction cost & 10 bps \\
\bottomrule
\end{tabular}
\end{adjustbox}
\caption{Training and inference parameters used in the main experiments.}
\label{tab:app_training_setup}
\end{table}

All reported training runs use \(4\times\) NVIDIA H800 GPUs. Under this setup,
training the CSI300 model takes around 36 hours, while training the CSI500
model takes around 60 hours.

\subsection{Baseline Descriptions}
\label{app:baselines}

The deterministic LLM baselines use the same daily agent interface, evaluator
tool, retained-pool updater, and backtesting protocol as \method, but without
GRPO training. We run \texttt{Qwen/Qwen3-4B-Instruct-2507},
\texttt{Qwen/Qwen3-30B-A3B-Instruct-2507}, and
\texttt{Qwen/Qwen3-235B-A22B-Instruct-2507} with temperature 0. These baselines
therefore test whether the base model alone can use the multi-turn interface
effectively before task-specific RL.

We do not include a reproduced AlphaAgent baseline in the main benchmark.
Although AlphaAgent is an important LLM-based prior method, the available
open-source release does not provide sufficiently detailed guidance for
faithful end-to-end replication in our setting, and we encountered multiple
implementation issues while trying to run the released code. We therefore
avoid reporting a potentially unfair or unstable reproduction result.

AlphaGen and AlphaForge are reproduced with their official implementations. For
AlphaGen, the reported native baseline uses the released factor-pool outputs
rather than direct stock-score files: we evaluate the daily expressions,
convert factor values to daily z-scored signals, and combine them with the
original AlphaGen weights. For AlphaForge, the main paper uses only the native
combination setting, where we reconstruct the daily raw linear score from the
released factor expressions and original weights. All reproduced outputs are
then evaluated under a common backtesting interface. For these stochastic
methods, the main paper reports averages over three independent runs.

The seed-factor linear-regression baseline is designed to separate the value of
the fixed seed library from the value of LLM-based search. This baseline
performs no model generation. For each trading day, it reuses the fixed seed-factor library from
the dataset, deduplicates expressions, evaluates each expression on the rolling
panel window, and applies the shared factor-combination rule described in
Appendix~\ref{app:pool_combination_details}. The combined signal is traded with the same
portfolio rule used elsewhere: buy the equal-weight top-\(K\) stocks and charge
10 bps transaction costs based on turnover. DeepSeek-R1 direct is a
direct-propose reference baseline: it proposes daily factor sets without tool
feedback, and the resulting factor sets are evaluated by the same
NNLS/ridge-style portfolio backtester. Table~\ref{tab:app_seed_factors} lists the fixed seed
factor library used both by this baseline and as the candidate manifest exposed
to the agent.

\subsection{Detailed Multi-Run Results}
\label{app:detailed_results}

Table~\ref{tab:app_main_results_detailed} expands the main comparison with run
counts, annualized returns, and run-to-run uncertainty. For \method, AlphaGen,
and AlphaForge, values are mean $\pm$ sample standard deviation over three
independent runs. The deterministic LLM baselines, DeepSeek-R1 direct, and the
seed-factor baseline are single-run references. Returns, annualized returns, and
drawdowns are percentages.
The matched single-run design ablations in
Table~\ref{tab:app_csi300_design_ablation} retain their original results and
are distinct from the three-run controlled comparison.

\begin{table*}[t]
\centering
\scriptsize
\setlength{\tabcolsep}{3pt}
\begin{adjustbox}{max width=\textwidth,center}
\begin{tabular}{llrrrrrrr}
\toprule
Universe & Method & Runs & Total Ret. & Ann. Ret. & Sharpe & Max DD & IC & RankIC \\
\midrule
CSI300 & \method & 3 & \(42.93 \pm 5.07\) & \(80.18 \pm 10.58\) & \(2.46 \pm 0.26\) & \(-9.34 \pm 0.68\) & \(0.0403 \pm 0.0065\) & \(0.0595 \pm 0.0115\) \\
& Qwen3-4B-Instruct-2507 & 1 & 0.58 & 0.95 & 0.16 & -16.42 & -0.0165 & -0.0133 \\
& Qwen3-30B-A3B-Instruct-2507 & 1 & -1.72 & -2.82 & -0.01 & -24.73 & 0.0030 & 0.0146 \\
& Qwen3-235B-A22B-Instruct-2507 & 1 & 6.88 & 11.59 & 0.55 & -16.32 & 0.0093 & 0.0226 \\
& DeepSeek-R1 direct & 1 & 22.79 & 40.24 & 1.21 & -16.20 & 0.0114 & -0.0033 \\
& Seed factors + LR & 1 & 34.51 & 62.96 & 1.82 & -18.65 & 0.0366 & 0.0361 \\
& AlphaGen & 3 & \(25.17 \pm 9.81\) & \(45.06 \pm 18.41\) & \(2.26 \pm 0.69\) & \(-8.24 \pm 0.39\) & \(-0.0009 \pm 0.0034\) & \(0.0109 \pm 0.0048\) \\
& AlphaForge & 3 & \(2.18 \pm 11.86\) & \(4.12 \pm 20.18\) & \(0.07 \pm 1.28\) & \(-12.15 \pm 0.39\) & \(-0.0086 \pm 0.0061\) & \(0.0024 \pm 0.0073\) \\
\midrule
CSI500 & \method & 3 & \(54.19 \pm 7.84\) & \(104.25 \pm 17.24\) & \(2.86 \pm 0.23\) & \(-14.34 \pm 2.45\) & \(0.0259 \pm 0.0038\) & \(0.0624 \pm 0.0034\) \\
& Qwen3-4B-Instruct-2507 & 1 & -4.83 & -7.84 & -0.30 & -21.60 & -0.0207 & 0.0090 \\
& Qwen3-30B-A3B-Instruct-2507 & 1 & 10.42 & 17.73 & 0.89 & -12.79 & 0.0064 & 0.0409 \\
& Qwen3-235B-A22B-Instruct-2507 & 1 & 7.10 & 11.96 & 0.54 & -16.17 & -0.0056 & 0.0064 \\
& DeepSeek-R1 direct & 1 & 14.07 & 24.22 & 0.84 & -18.60 & 0.0042 & -0.0211 \\
& Seed factors + LR & 1 & 18.56 & 32.36 & 1.07 & -22.67 & 0.0201 & 0.0189 \\
& AlphaGen & 3 & \(48.85 \pm 17.49\) & \(93.47 \pm 37.87\) & \(2.82 \pm 0.33\) & \(-13.45 \pm 3.07\) & \(0.0060 \pm 0.0101\) & \(0.0042 \pm 0.0091\) \\
& AlphaForge & 3 & \(36.32 \pm 33.93\) & \(70.19 \pm 71.29\) & \(1.97 \pm 0.67\) & \(-11.11 \pm 4.68\) & \(-0.0091 \pm 0.0089\) & \(0.0086 \pm 0.0135\) \\
\bottomrule
\end{tabular}
\end{adjustbox}
\caption{Detailed out-of-sample results corresponding to main-paper
Tables~\ref{tab:csi300_main} and~\ref{tab:csi500_main}.}
\label{tab:app_main_results_detailed}
\end{table*}

Table~\ref{tab:app_benchmark_relative} contextualizes portfolio performance
against broad market movement using an equal-weight benchmark constructed from the same
tradable universe over the same test period. Daily excess returns are portfolio
returns minus benchmark returns; cumulative excess return is the difference
between cumulative portfolio and benchmark returns. The table also reports the
Sharpe ratio of the daily excess-return series for the three stochastic
factor-mining methods.

\begin{table*}[t]
\centering
\setlength{\tabcolsep}{4pt}
\begin{adjustbox}{max width=0.9\textwidth,center}
\begin{tabular}{llrrrr}
\toprule
Universe & Method & Portfolio Ret. & Benchmark Ret. & Excess Ret. & Excess Sharpe \\
\midrule
CSI300 & \method & $42.93 \pm 5.07$ & 21.44 & \textbf{$21.49 \pm 5.07$} & \textbf{$1.65 \pm 0.29$} \\
& AlphaGen & $25.17 \pm 9.81$ & 21.44 & $3.73 \pm 9.81$ & $0.32 \pm 1.34$ \\
& AlphaForge & $2.18 \pm 11.86$ & 21.44 & $-19.26 \pm 11.86$ & $-2.39 \pm 1.47$ \\
\midrule
CSI500 & \method & $54.19 \pm 7.84$ & 35.97 & \textbf{$18.22 \pm 7.84$} & \textbf{$1.36 \pm 0.43$} \\
& AlphaGen & $48.85 \pm 17.49$ & 35.97 & $12.88 \pm 17.49$ & $0.95 \pm 1.16$ \\
& AlphaForge & $36.32 \pm 33.93$ & 35.97 & $0.35 \pm 33.93$ & $-1.00 \pm 2.56$ \\
\bottomrule
\end{tabular}
\end{adjustbox}
\caption{Benchmark-relative out-of-sample performance. Values for each method
are mean $\pm$ sample standard deviation over three independent runs. Returns
are percentages; bold marks the best excess result in each universe.}
\label{tab:app_benchmark_relative}
\end{table*}

\subsection{Ablation Design}
\label{app:ablation_design}

Table~\ref{tab:csi300_ablation} in the main paper separates the controlled
multi-turn comparison from the matched single-run design ablations.
The \textit{full method} is our main configuration: it uses multi-turn tool use,
single-factor evaluator feedback, retained-pool updates with auxiliary
regularization terms, and memory feedback about recently tried factors and the
current pool. The \textit{full + reward auxiliary} variant keeps the same
rollout and pool-updater design, but additionally adds the pool regularization
terms to the trajectory-level reward computation just like the pool score. This
tests whether applying diversity, stability, and redundancy preferences
directly in the RL reward improves learning beyond using them inside pool
selection.

The \textit{no pool auxiliary score} variant removes the auxiliary
regularization terms from the retained-pool score, so pool updates rely on
predictive metrics without the additional diversity, stability, and redundancy
preferences. The \textit{no memory feedback} variant keeps the evaluator and
pool updater unchanged, but hides the two extra memory channels from the agent:
the recent history of tried factors and the current retained-pool summary. This
tests whether multi-turn search benefits from remembering both failed
candidates and the evolving pool state.

The controlled \textit{GRPO single-turn} baseline instead retains the same
Qwen3-4B backbone, GRPO training algorithm, data splits, and terminal
pool-level reward as \method, but directly generates the final factor pool in
one response. Its three-run results are reported in
Table~\ref{tab:app_csi300_ablation_detailed}; this control separates the effect of
multi-turn tool-based refinement from the effect of GRPO training itself.

\begin{table*}[t]
\centering
\begin{subtable}[t]{\textwidth}
\centering
\scriptsize
\setlength{\tabcolsep}{3pt}
\begin{adjustbox}{max width=\textwidth,center}
\begin{tabular}{lrrrrrrr}
\toprule
Method & Runs & Total Ret. & Ann. Ret. & Sharpe & Max DD & IC & RankIC \\
\midrule
\method & 3 & $42.93 \pm 5.07$ & $80.18 \pm 10.58$ & $2.46 \pm 0.26$ & $-9.34 \pm 0.68$ & $0.0403 \pm 0.0065$ & $0.0595 \pm 0.0115$ \\
GRPO single-turn & 3 & $35.62 \pm 9.53$ & $64.29 \pm 20.84$ & $2.02 \pm 0.45$ & $-16.32 \pm 4.50$ & $0.0407 \pm 0.0103$ & $0.0496 \pm 0.0127$ \\
\bottomrule
\end{tabular}
\end{adjustbox}
\caption{Controlled multi-turn comparison. Values are mean $\pm$ sample
standard deviation over three independent runs.}
\label{tab:app_csi300_turn_control}
\end{subtable}

\vspace{0.6em}
\begin{subtable}[t]{\textwidth}
\centering
\setlength{\tabcolsep}{4pt}
\begin{adjustbox}{max width=0.9\textwidth,center}
\begin{tabular}{lrrrrrrr}
\toprule
Variant & Avg. Turns & Total Ret. & Ann. Ret. & Sharpe & Max DD & IC & RankIC \\
\midrule
Full method & 5.24 & 48.64 & 92.09 & 2.74 & -10.00 & 0.0361 & 0.0501 \\
Full + reward auxiliary & 5.61 & 44.49 & 83.34 & 2.50 & -10.00 & 0.0408 & 0.0594 \\
No pool auxiliary score & 5.67 & 34.39 & 62.72 & 2.06 & -8.53 & 0.0419 & 0.0634 \\
No memory feedback & 15.75 & 26.40 & 47.09 & 1.70 & -10.99 & 0.0293 & 0.0473 \\
\bottomrule
\end{tabular}
\end{adjustbox}
\caption{Matched single-run design ablations. Avg. Turns is the average number
of assistant turns per daily rollout.}
\label{tab:app_csi300_design_ablation}
\end{subtable}
\caption{Detailed CSI300 controlled comparisons corresponding to
Table~\ref{tab:csi300_ablation} in the main paper. Returns and drawdowns are
percentages.}
\label{tab:app_csi300_ablation_detailed}
\end{table*}

\begin{table*}[t]
\centering
\small
\renewcommand{\arraystretch}{1.12}
\setlength{\tabcolsep}{6pt}
\begin{adjustbox}{max width=\textwidth,center}
\begin{tabular}{lp{0.78\textwidth}}
\toprule
Family & Expression \\
\midrule
Reversal & \texttt{-rank(ts\_delta(close, 1))} \\
Reversal & \texttt{-rank(close / delay(close, 5) - 1)} \\
Momentum & \texttt{rank(close / delay(close, 10) - 1)} \\
Momentum & \texttt{rank(close / delay(close, 20) - 1)} \\
Trend & \texttt{rank(close / ts\_mean(close, 20) - 1)} \\
Trend & \texttt{rank(close / ts\_mean(close, 50) - 1)} \\
Trend & \texttt{-rank(close / ts\_mean(close, 20) - 1)} \\
Volume & \texttt{rank(volume / ts\_mean(volume, 20))} \\
Volume & \texttt{-rank(volume / ts\_mean(volume, 20))} \\
Price-volume & \texttt{rank(ts\_corr(close, volume, 10))} \\
Price-volume & \texttt{-rank(ts\_corr(close, volume, 10))} \\
Price-volume & \texttt{rank(ts\_corr(close, volume, 5))} \\
Vol.-adjusted & \texttt{rank((close / delay(close, 5) - 1) / ts\_std(ret\_1d, 20))} \\
Vol.-adjusted & \texttt{rank(close / delay(close, 10) - 1) - rank(ts\_std(ret\_1d, 20))} \\
Vol.-adjusted & \texttt{rank(close / ts\_mean(close, 20) - 1) - rank(ts\_std(ret\_1d, 20))} \\
Hybrid & \texttt{rank(ts\_corr(close, volume, 10)) - rank(ts\_std(ret\_1d, 20))} \\
Hybrid & \texttt{rank(close / delay(close, 10) - 1) + rank(volume / ts\_mean(volume, 20))} \\
Hybrid & \texttt{rank(close / ts\_mean(close, 20) - 1) + rank(ts\_corr(close, volume, 10))} \\
VWAP & \texttt{rank(vwap\_spread)} \\
VWAP & \texttt{-rank(vwap\_spread)} \\
\bottomrule
\end{tabular}
\end{adjustbox}
\caption{Fixed seed factor library used by the seed-factor baseline and as the
candidate manifest for agent inference.}
\label{tab:app_seed_factors}
\end{table*}

\section{Efficiency, Pool Behavior, and Training Dynamics}
\label{app:efficiency_training}

This section collects additional empirical evidence that supports the main
results from three angles: runtime cost, retained-pool behavior, and learning
dynamics. The runtime tables show the daily cost of representative LLM-based
inference settings; the pool-analysis subsections summarize how large the
retained pool tends to be and how closely it tracks the seed library; and the
training-dynamics plot illustrates how the GRPO policy changes during
optimization. The goal is not to claim hardware-independent latency, but to
show the practical cost and search behavior of using a small trained policy
with evaluator access compared with larger LLM-only inference settings under
the same daily interface.

\subsection{Inference Efficiency}
\label{app:inference_efficiency}

Tables~\ref{tab:app_inference_efficiency_csi300}
and~\ref{tab:app_inference_efficiency_csi500} summarize per-day runtime for
representative LLM-based methods on CSI300 and CSI500. For \method and the
deterministic Qwen baselines, the reported numbers come from logged day-by-day
inference-time metadata. For DeepSeek-R1 direct, we report the final average
per-day cost after adding one-shot generation time for the full test horizon to
the logged backtesting time and dividing by the number of trading days.

\begin{table}[t]
\centering
\setlength{\tabcolsep}{4pt}
\begin{adjustbox}{max width=\linewidth}
\begin{tabular}{lr}
\toprule
Method & Sec./day \\
\midrule
\method & 100.62 \\
Qwen3-30B-A3B-Instruct-2507 & 92.07 \\
Qwen3-235B-A22B-Instruct-2507 & 377.53 \\
DeepSeek-R1 one-shot & 248.98 \\
\bottomrule
\end{tabular}
\end{adjustbox}
\caption{Per-day runtime on CSI300. The ``Sec./day'' column reports average
cost per trading day. For DeepSeek-R1 direct, the value includes both one-shot
generation over the test horizon and the subsequent backtesting time.}
\label{tab:app_inference_efficiency_csi300}
\end{table}

\begin{table}[t]
\centering
\setlength{\tabcolsep}{4pt}
\begin{adjustbox}{max width=\linewidth}
\begin{tabular}{lr}
\toprule
Method & Sec./day \\
\midrule
\method & 232.54 \\
Qwen3-30B-A3B-Instruct-2507 & 182.94 \\
Qwen3-235B-A22B-Instruct-2507 & 527.99 \\
DeepSeek-R1 one-shot & 254.82 \\
\bottomrule
\end{tabular}
\end{adjustbox}
\caption{Per-day runtime on CSI500. The ``Sec./day'' column reports average
cost per trading day. For DeepSeek-R1 direct, the value includes both one-shot
generation over the test horizon and the subsequent backtesting time.}
\label{tab:app_inference_efficiency_csi500}
\end{table}

The timing results show two useful patterns. First, \method remains much faster
than the 235B LLM setting on both universes while still using multi-turn tool
feedback, and it is also more practical than the final end-to-end cost of the
DeepSeek-R1 direct pipeline once one-shot generation is included.
Second, the larger CSI500 universe increases the per-day evaluation cost for
every LLM-based setting. This supports our design choice: train a small policy
to use data-access tools effectively, rather than relying only on a much larger
reasoning model at inference time.

\subsection{GRPO Training Dynamics}
\label{app:training_dynamics}

\begin{figure*}[t]
\centering
\includegraphics[width=0.78\textwidth]{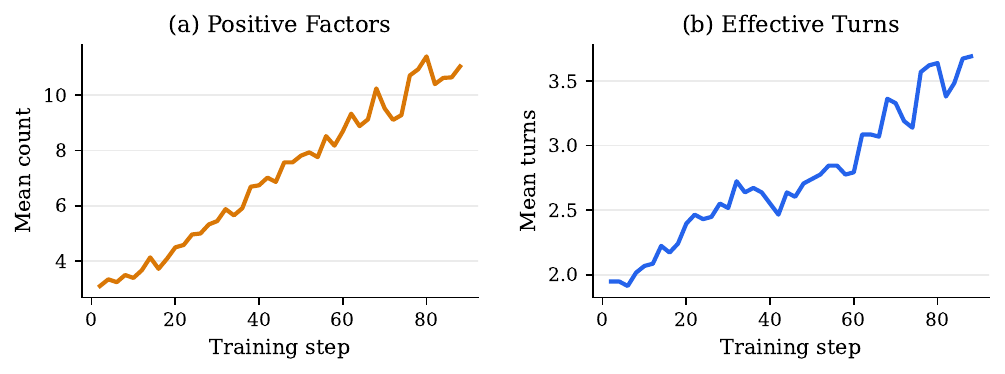}
\caption{Training dynamics for the best CSI300 \method checkpoint. The left
panel shows the mean number of positive-scoring submitted factors per rollout,
and the right panel shows the mean effective turn count during GRPO training.}
\label{fig:app_training_dynamics}
\end{figure*}

Figure~\ref{fig:app_training_dynamics} visualizes two training metrics from the
best CSI300 checkpoint: the mean number of positive-scoring submitted factors
per rollout and the mean effective turn count. Both trends improve over
training, suggesting that the policy learns to keep searching for longer and to
propose a larger set of usable factors.

The training curves are consistent with the qualitative case studies: after
GRPO, the policy does not simply stop after a plausible first proposal. It makes
more effective tool calls, receives more usable feedback, and builds richer
candidate pools before the reward is computed.

\subsection{Average Daily Pool Size}
\label{app:avg_pool_size}

\begin{table}[t]
\centering
\setlength{\tabcolsep}{4pt}
\begin{adjustbox}{max width=\linewidth}
\begin{tabular}{lrr}
\toprule
Method & CSI300 & CSI500 \\
\midrule
\method & 6.82 & 8.23 \\
Qwen3-4B-Instruct-2507 & 2.10 & 1.67 \\
Qwen3-30B-A3B-Instruct-2507 & 3.28 & 2.78 \\
Qwen3-235B-A22B-Instruct-2507 & 7.29 & 7.14 \\
DeepSeek-R1 direct & 15.95 & 16.41 \\
Seed factors + LR & 20.00 & 20.00 \\
AlphaGen & 20.00 & 20.00 \\
AlphaForge & 10.00 & 10.00 \\
\bottomrule
\end{tabular}
\end{adjustbox}
\caption{Average daily pool size for the main comparison methods. For
stochastic baselines such as AlphaGen and AlphaForge, we report the same
three-run average used in the main tables.}
\label{tab:app_avg_pool_size}
\end{table}

Table~\ref{tab:app_avg_pool_size} shows that \method maintains a relatively
reasonable retained-pool size on both universes. Its average pool size is 6.82
on CSI300 and 8.23 on CSI500, which is larger than the weakest deterministic
LLM baselines but still clearly smaller than the fixed-pool seed baseline,
AlphaGen, AlphaForge, and DeepSeek-R1 direct. This pattern is consistent with
the design of \method: rather than accumulating as many factors as possible, it
keeps a moderate adaptive pool and updates it according to marginal pool-level
utility. The result is a pool that is neither trivially small nor inflated by
redundant factors, which helps explain why \method can achieve strong
portfolio-level performance without relying on unusually large retained sets.

\subsection{Seed-Pool Similarity Analysis}
\label{app:seed_pool_similarity}

\begin{table}[t]
\centering
\setlength{\tabcolsep}{4pt}
\begin{adjustbox}{max width=\linewidth}
\begin{tabular}{lcc}
\toprule
Universe & Avg. exact & Unique exact \\
\midrule
CSI300 & 0.5069 & 0.0380 \\
CSI500 & 0.4851 & 0.0315 \\
\bottomrule
\end{tabular}
\end{adjustbox}
\caption{Seed-pool similarity statistics for \method. ``Avg. exact'' is the
average daily fraction of retained factors that are exact seed formulas, and
``Unique exact'' is the fraction of unique final expressions over the full run
that exactly match the seed library.}
\label{tab:app_seed_pool_similarity}
\end{table}

Table~\ref{tab:app_seed_pool_similarity} shows that the retained pool is not a
trivial copy of the seed library. Although about half of the active factors on
any given day are exact seed formulas, the run-level exact-overlap ratio over
unique discovered expressions is only 3--4\% in both universes. This means the
agent does produce many non-seed formulas over time rather than repeatedly
cycling through the initialization set.

Taken together, these two overlap views suggest that the search remains
meaningfully seed-anchored at the daily level while still expanding beyond the
initialization library over the full run. In other words, \method does not
simply recycle the seed pool; it retains some strong seed anchors but also
introduces many new formulas over time.

\section{Case-Study Submission Trace}
\label{app:case_submission_trace}

\subsection{Complete Submission Trace}
\label{app:complete_submission_trace}

Table~\ref{tab:app_case_submissions} lists all evaluated factor submissions for
the CSI300 \method rollout on the first formal test trading day. The main text
reports only accepted additions; this table also includes rejected submissions
to show how turn-level tool feedback redirects the search.

The trace illustrates how the data-accessible tool changes the search direction
within a single daily rollout. The natural-language regime description for
2025-07-11 is supportive: long-term momentum is strong, medium-term trend is
healthy, and short-term liquidity is positive. The agent therefore begins with
momentum and trend hypotheses, but the first four submissions have weak or
negative RankIC and are immediately rejected. The first accepted factors are
instead short-horizon reversal signals, which raise the pool score to 0.0661
after two additions.

The next phase shows the role of retained-pool feedback. Some candidates have
reasonable standalone scores but do not help the pool. For example, the two
hybrid seed-style factors in rows 7--8 have positive IC/RankIC, yet both have
negative marginal gain and are rejected. Later, volatility-adjusted variants in
rows 17--19 achieve even stronger standalone IC/RankIC, including RankIC 0.0689
for \texttt{-rank(vwap\_spread) - rank(ts\_std(ret\_1d, 20))}, but they are also
rejected because they reduce the retained-pool objective. This is the key
difference between screening factors independently and optimizing an evolving
pool: the accepted set is determined by incremental contribution, not by
single-factor score alone.

The final phase expands the pool with structurally different liquidity-reversal
signals. The agent attempts two invalid VWAP lookback variants, receives explicit
invalid-expression feedback, and then moves to valid alternatives such as
\texttt{-rank(close / volume)}, volume-delta reversals, and longer
price-volume-correlation reversals. These factors increase the pool score from
0.0813 to 0.0938 and produce a final pool of 14 retained factors. The final
pool is coherent but not redundant: it combines VWAP reversal, short-horizon
price reversal, volume-momentum reversal, and price-volume-correlation reversal
across several lookbacks.

Overall, this rollout highlights four desirable properties of \method. First,
the policy is willing to revise its initial regime-consistent hypothesis once
the evaluator provides contradictory evidence. Second, it optimizes for pool
quality rather than standalone factor score, rejecting several individually
strong but redundant candidates. Third, the retained-pool memory makes the
later search progressively more structured, with new proposals built as local
refinements around accepted anchors instead of repeated blind exploration.
Finally, the resulting pool is both diverse and interpretable: the accepted
factors cluster around a small number of coherent economic motifs while still
covering multiple lookbacks and liquidity-sensitive structures.

\section{Prompt and Feedback Templates}
\label{app:prompts}

We include the prompts needed to reproduce the language interface. The
multi-turn prompt is used by \method during training and inference, and the
verbalization prompt converts numerical market-state summaries into the regime
descriptions seen by the agent. Dynamic fields in braces are filled from each
rolling decision example, so the operator list, seed-factor manifest, and date
windows are instance-specific.

\subsection{Multi-Turn Tool-Using Prompt}
\label{app:multi_turn_prompt}

The multi-turn system prompt defines the evaluator tool, the valid factor
families, the DSL constraints, and the behavioral rules for iterative search.
It is intentionally explicit because the policy must learn to treat evaluator
feedback and retained-pool memory as part of the action context.

\begin{figure*}[p]
    \centering
    \includegraphics[width=\textwidth,keepaspectratio]{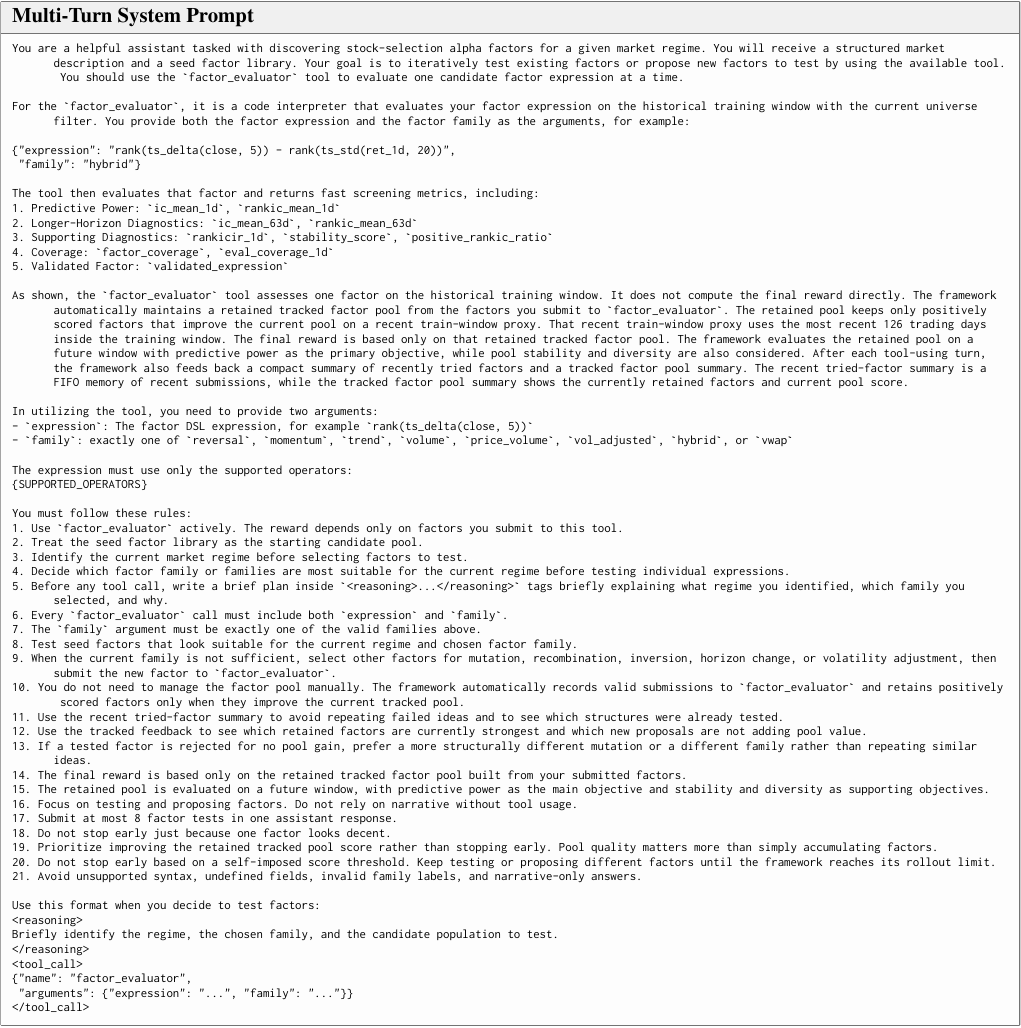}
    \caption{Multi-turn system prompt used by
    \method.}
    \label{fig:prompt_multi_system}
\end{figure*}

\begin{figure*}[p]
    \centering
    \includegraphics[width=\textwidth,keepaspectratio]{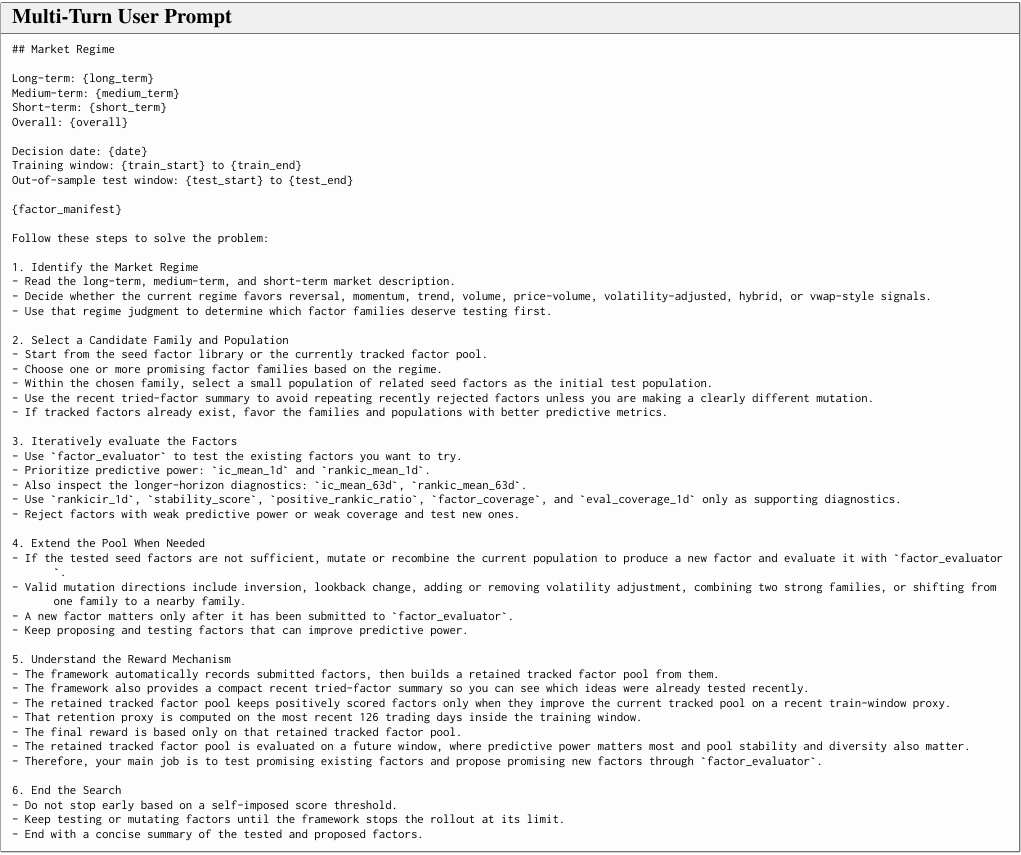}
    \caption{Multi-turn user prompt for a single
    decision date.}
    \label{fig:prompt_multi_user}
\end{figure*}

The user prompt instantiates the daily decision problem. It inserts the
market-regime description, rolling train/test windows, seed factor manifest,
and the step-by-step search protocol used by the agent.
Figures~\ref{fig:prompt_multi_system} and~\ref{fig:prompt_multi_user} show
these two prompt components.

\subsection{Example Market Description}
\label{app:example_market_description}

The agent receives market context as concise language rather than raw numerical
tables. Each description compresses the structured short-, medium-, and
long-horizon market summaries into a compact prompt context that can be read by
the language policy before factor search begins. Table~\ref{tab:app_market_description_block}
shows the description used for the first formal CSI300 test day in the main
\method rollout.

\begin{table}[H]
\centering
\small
\setlength{\tabcolsep}{4pt}
\begin{adjustbox}{max width=\linewidth}
\begin{tabular}{lp{0.72\linewidth}}
\toprule
Horizon & Description \\
\midrule
Long-term & The market is in a strong long-term uptrend with moderate
dispersion and stable volatility. \\
Medium-term & The medium-term trend remains positive with moderate
breadth and increasing trading activity. \\
Short-term & Recent trading shows mild positive momentum with balanced
participation, strong liquidity, and leadership from the Postal Service sector. \\
Overall & The market environment is supportive with sustained long-term
momentum, healthy medium-term trends, and positive short-term dynamics. \\
\bottomrule
\end{tabular}
\end{adjustbox}
\caption{Example natural-language market description supplied to the agent.}
\label{tab:app_market_description_block}
\end{table}

\subsection{Market-State Verbalization Prompt}
\label{app:market_verbalization_prompt}

The market-state verbalization prompt is used before agent inference. It maps
structured numerical summaries into concise long-, medium-, short-term, and
overall regime descriptions, which are then inserted into the multi-turn user
prompt.

\begin{figure*}[p]
    \centering
    \includegraphics[width=\textwidth,keepaspectratio]{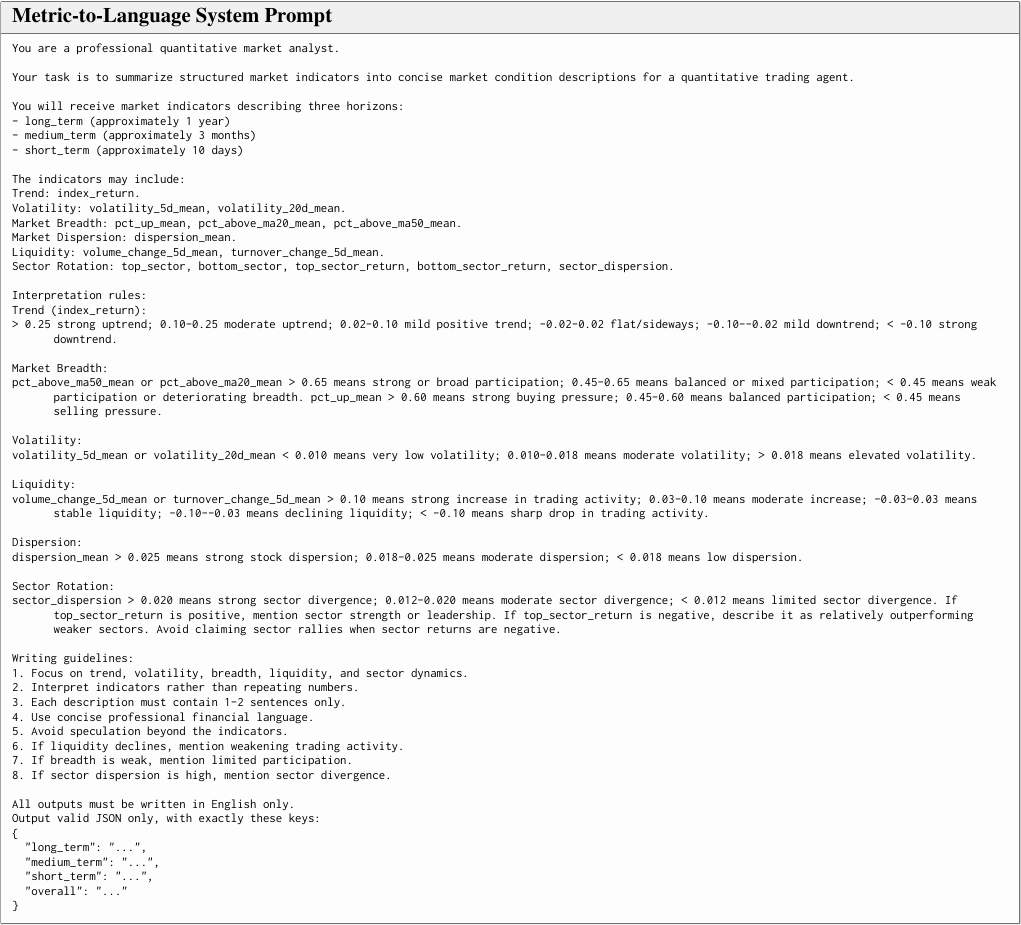}
    \caption{Market-state verbalization system
    prompt.}
    \label{fig:prompt_metric_system}
\end{figure*}

\begin{figure*}[p]
    \centering
    \includegraphics[width=\textwidth,keepaspectratio]{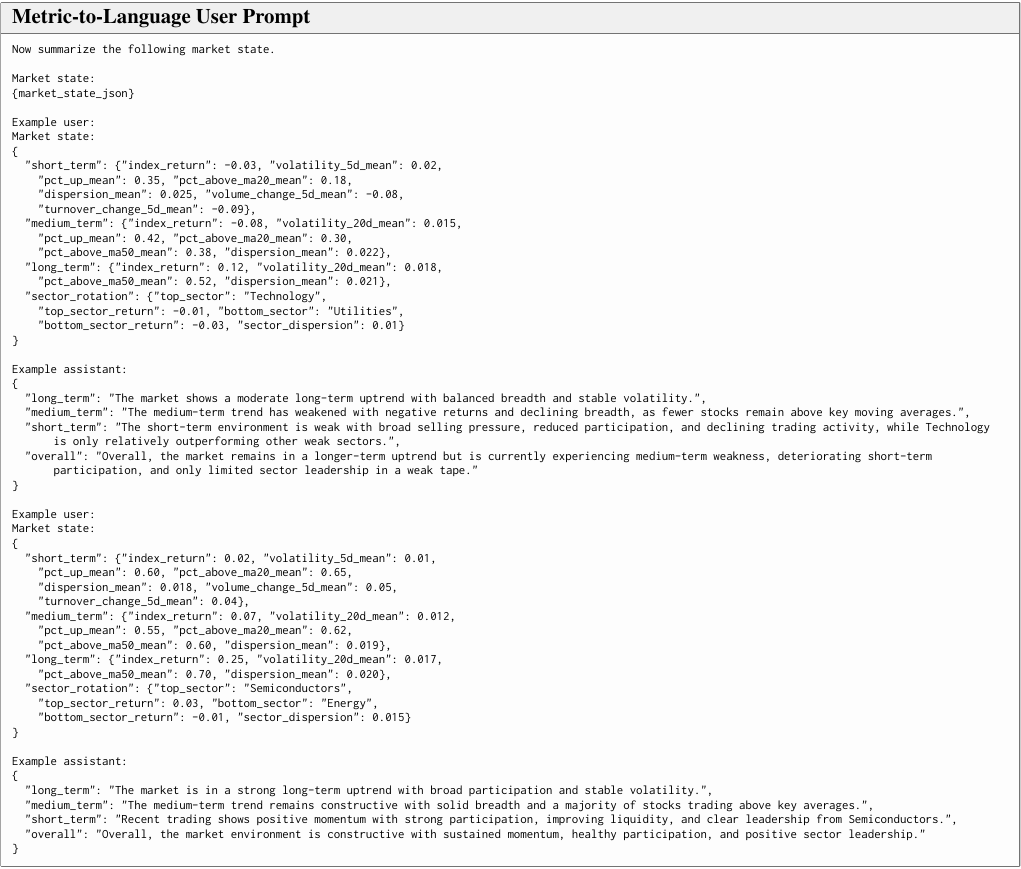}
    \caption{Market-state verbalization user
    prompt with few-shot examples.}
    \label{fig:prompt_metric_user}
\end{figure*}

The metric-to-language user prompt supplies the structured market-state object
for a single decision date. We also include the few-shot messages used to make
the wording concise and consistent across dates.
Figures~\ref{fig:prompt_metric_system} and~\ref{fig:prompt_metric_user} show
the system and user prompts used in this verbalization stage.

\subsection{Memory Feedback Template}
\label{app:tool_feedback_prompt}

\begin{figure*}[p]
    \centering
    \includegraphics[width=\textwidth,keepaspectratio]{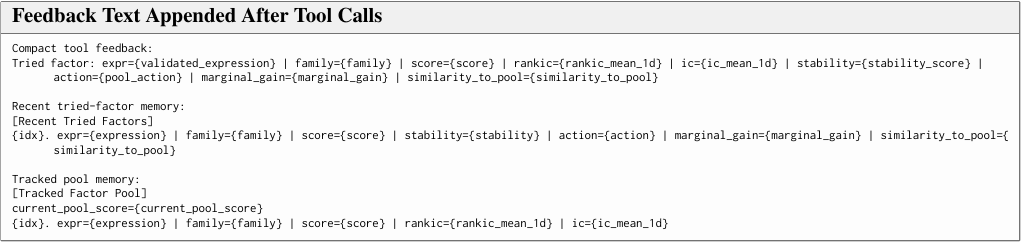}
    \caption{Feedback template appended after each tool call.}
    \label{fig:prompt_feedback}
\end{figure*}

The feedback template is appended after tool calls. It gives the agent three
compact memory channels: the current factor's single-factor metrics and pool
decision, a FIFO list of recently tried factors, and the current retained-pool
state. This is the main interface through which evaluator data enters later
turns. Figure~\ref{fig:prompt_feedback} shows the exact feedback block appended
at the end of each tool-using turn.

\begin{table*}[p]
\centering
\footnotesize
\renewcommand{\arraystretch}{1.02}
\setlength{\tabcolsep}{2pt}
\begin{adjustbox}{width=\textwidth,center}
\begin{tabular}{@{}ccp{0.61\textwidth}rrrrl@{}}
\toprule
\# & T & Expression & Score & IC & RankIC & Pool & Decision \\
\midrule
1 & 1 & \texttt{rank(close / delay(close, 10) - 1)} & 0.000 & 0.00207 & -0.0110 & -- & reject \\
2 & 1 & \texttt{rank(close / delay(close, 20) - 1)} & 0.000 & -0.00114 & -0.0134 & -- & reject \\
3 & 1 & \texttt{rank(close / ts\_mean(close, 20) - 1)} & 0.000 & 0.00111 & -0.0130 & -- & reject \\
4 & 1 & \texttt{rank(close / ts\_mean(close, 50) - 1)} & 0.000 & 0.00187 & -0.00816 & -- & reject \\
5 & 1 & \texttt{-rank(ts\_delta(close, 1))} & 0.380 & 0.0106 & 0.0213 & 0.0432 & accept \\
6 & 1 & \texttt{-rank(close / delay(close, 5) - 1)} & 0.171 & 0.000166 & 0.0136 & 0.0661 & accept \\
7 & 1 & \texttt{rank(ts\_corr(close, volume, 10)) - rank(ts\_std(ret\_1d, 20))} & 0.505 & 0.0146 & 0.0299 & 0.0661 & reject \\
8 & 1 & \texttt{rank(ts\_corr(close, volume, 5)) - rank(ts\_std(ret\_1d, 20))} & 0.592 & 0.0194 & 0.0351 & 0.0661 & reject \\
9 & 2 & \texttt{-rank(ts\_delta(close, 3))} & 0.190 & 0.00218 & 0.0132 & 0.0665 & accept \\
10 & 2 & \texttt{-rank(ts\_delta(close, 7))} & 0.0329 & -0.00388 & 0.00651 & 0.0674 & accept \\
11 & 2 & \texttt{-rank(ts\_delta(close, 15))} & 0.0494 & -0.00253 & 0.00649 & 0.0674 & accept \\
12 & 2 & \texttt{-rank(close / delay(close, 10) - 1)} & 0.111 & -0.00207 & 0.0110 & 0.0674 & reject \\
13 & 2 & \texttt{-rank(close / delay(close, 20) - 1)} & 0.180 & 0.00114 & 0.0134 & 0.0674 & reject \\
14 & 2 & \texttt{-rank(ts\_corr(close, volume, 10))} & 0.355 & 0.00826 & 0.0214 & 0.0727 & accept \\
15 & 2 & \texttt{-rank(ts\_corr(close, volume, 5))} & 0.136 & 0.000004 & 0.0110 & 0.0727 & accept \\
16 & 2 & \texttt{-rank(vwap\_spread)} & 0.637 & 0.0249 & 0.0353 & 0.0813 & accept \\
17 & 3 & \texttt{-rank(ts\_delta(close, 1)) - rank(ts\_std(ret\_1d, 20))} & 0.790 & 0.0271 & 0.0587 & 0.0813 & reject \\
18 & 3 & \texttt{-rank(ts\_corr(close, volume, 10)) - rank(ts\_std(ret\_1d, 20))} & 0.740 & 0.0225 & 0.0536 & 0.0813 & reject \\
19 & 3 & \texttt{-rank(vwap\_spread) - rank(ts\_std(ret\_1d, 20))} & 0.872 & 0.0385 & 0.0689 & 0.0813 & reject \\
20 & 3 & \texttt{-rank(ts\_delta(close, 1)) + rank(ts\_corr(close, volume, 10))} & 0.0232 & 0.00204 & -0.000183 & 0.0813 & reject \\
21 & 3 & \texttt{-rank(vwap\_spread) + rank(ts\_corr(close, volume, 10))} & 0.262 & 0.0114 & 0.00998 & 0.0813 & reject \\
22 & 3 & \texttt{-rank(ts\_delta(close, 1)) * rank(ts\_corr(close, volume, 10))} & 0.411 & 0.0131 & 0.0218 & 0.0813 & reject \\
23 & 3 & \texttt{-rank(vwap\_spread) * rank(ts\_corr(close, volume, 10))} & 0.566 & 0.0217 & 0.0297 & 0.0813 & reject \\
24 & 4 & \texttt{rank(ts\_delta(close, 1))} & 0.000 & -0.0106 & -0.0213 & 0.0813 & reject \\
25 & 4 & \texttt{-rank(close / volume)} & 0.473 & 0.0157 & 0.0254 & 0.0844 & accept \\
26 & 4 & \texttt{-rank(ts\_delta(volume, 5))} & 0.336 & 0.0103 & 0.0176 & 0.0885 & accept \\
27 & 4 & \texttt{-rank(ts\_delta(volume, 10))} & 0.386 & 0.0116 & 0.0209 & 0.0885 & accept \\
28 & 4 & \texttt{-rank(ts\_delta(close, 30))} & 0.0834 & -0.000385 & 0.00707 & 0.0917 & accept \\
29 & 4 & \texttt{-rank(ts\_corr(close, volume, 20))} & 0.270 & 0.00626 & 0.0159 & 0.0931 & accept \\
30 & 4 & \texttt{-rank(ts\_corr(close, volume, 30))} & 0.153 & 0.00333 & 0.00904 & 0.0938 & accept \\
\bottomrule
\end{tabular}
\end{adjustbox}
\caption{Complete evaluated factor submissions for the CSI300 \method case
study on 2025-07-11. The Pool column reports the retained-pool score after the
feedback event; ``--'' indicates that no factor had yet entered the retained
pool. The table omits factor-family labels and marginal pool gain to keep the
turn-level trace readable.}
\label{tab:app_case_submissions}
\end{table*}

\end{document}